\documentclass[manuscript,screen]{acmart}

\usepackage{amsmath}
\usepackage{epsfig,graphicx,float}
\usepackage[caption=false]{subfig}
\usepackage{booktabs,multirow,array} 

\usepackage[misc]{ifsym}
\usepackage[figuresright]{rotating}

\usepackage{color}
\usepackage{wrapfig}

\usepackage{makecell}

\usepackage{xcolor}
\usepackage{hyperref}
\hypersetup{colorlinks=true}

\setcopyright{acmcopyright}
\copyrightyear{2018}
\acmYear{2018}
\acmDOI{XXXXXXX.XXXXXXX}

\begin{document}

\title {Empowering Agrifood System with Artificial Intelligence: A Survey of the Progress, Challenges and Opportunities}

\author{Tao Chen}\authornote{The first three authors contributed equally to this work.}
\author{Liang Lv}\authornotemark[1]
\affiliation{%
  \institution{China University of Geosciences}
  \city{Wuhan}
  \country{China}
}
\email{{taochen,lvl14}@cug.edu.cn}

\author{Di Wang}\authornotemark[1]
\affiliation{%
  \institution{Wuhan University}
  \city{Wuhan}
  \country{China}}
\email{wd74108520@gmail.com}

\author{Jing Zhang}\authornote{Corresponding authors}
\affiliation{%
  \institution{The University of Sydney}
  \city{Sydney}
  \country{Australia}}
\affiliation{%
  \institution{Wuhan University}
  \city{Wuhan}
  \country{China}}
\email{jingzhang.cv@gmail.com}

\author{Yue Yang}
\author{Zeyang Zhao}
\author{Chen Wang}
\author{Xiaowei Guo}
\author{Hao Chen}
\author{Qingye Wang}
\affiliation{%
  \institution{China University of Geosciences}
  \city{Wuhan}
  \country{China}
}

\author{Yufei Xu}
\author{Qiming Zhang}
\affiliation{%
  \institution{The University of Sydney}
  \city{Sydney}
  \country{Australia}}

\author{Bo Du}\authornotemark[2]
\author{Liangpei Zhang}
\email{{dubo,zlp62}@whu.edu.cn}
\affiliation{%
  \institution{Wuhan University}
  \city{Wuhan}
  \country{China}}

\author{Dacheng Tao}
\affiliation{%
  \institution{Nanyang Technological University}
  \country{Singapore}}
\email{dacheng.tao@ntu.edu.sg}

\renewcommand{\shortauthors}{Chen and Lv, et al.}

\begin{abstract}
 With the world population rapidly increasing, transforming our agrifood systems to be more productive, efficient, safe, and sustainable is crucial to mitigate potential food shortages. Recently, artificial intelligence (AI) techniques such as deep learning (DL) have demonstrated their strong abilities in various areas, including language, vision, remote sensing (RS), and agrifood systems applications. However, the overall impact of AI on agrifood systems remains unclear. In this paper, we thoroughly review how AI techniques can transform agrifood systems and contribute to the modern agrifood industry. Firstly, we summarize the data acquisition methods in agrifood systems, including acquisition, storage, and processing techniques. Secondly, we present a progress review of AI methods in agrifood systems, specifically in agriculture, animal husbandry, and fishery, covering topics such as agrifood classification, growth monitoring, yield prediction, and quality assessment. Furthermore, we highlight potential challenges and promising research opportunities for transforming modern agrifood systems with AI. We hope this survey could offer an overall picture to newcomers in the field and serve as a starting point for their further research. The project website is \url{https://github.com/Frenkie14/Agrifood-Survey}.
\end{abstract}



\maketitle

\section{Introduction}

The agrifood system encompasses a wide range of topics, such as agriculture, animal husbandry, and fisheries~\cite{liu2019estimating,ottinger2017large,zeng2019extracting}. With the world's population growing rapidly, there is an increased need for more effective agrifood systems to support billions of jobs and feed the global population. To this end, it is urgent to enhance food security and improve crop yield through dynamic monitoring of growth situations, optimized harvesting schedules, and reduced waste in current agrifood systems. Achieving this requires a smarter system that can handle the large amounts of data generated by agrifood systems and make accurate predictions.
Recent advances in artificial intelligence (AI) techniques have demonstrated their ability to handle large-scale data in various fields~\cite{dosovitskiyimage,xu2021vitae, wang_rsp_2022,wang2022advancing}, including natural language processing, computer vision, medical imaging, remote sensing (RS), and more. Different AI techniques have been explored and have shown promise in building stronger and smarter agrifood systems from various perspectives. For instance, AI can monitor agriculture growth by jointly analyzing temperature, soil, water, and gas conditions~\cite{shin2013development}, help distinguish crop diseases with hyperspectral data~\cite{dopper2022estimating}, predict yield through RS techniques~\cite{khanal2018integration}, monitor pasture~\cite{de2021predicting}, and identify fishing areas~\cite{zeng2019extracting}. Overall, incorporating AI techniques can lead to higher efficiency in agrifood systems with comprehensive analysis of the crops or animals' situation. Such a smart system can enhance food safety and better supply chain management by providing early detection of crop diseases and predicting yield.

Although there are many benefits to using AI techniques in agrifood systems, it comes with several challenges that need to be resolved for better agrifood systems. The first challenge is collecting potential data that can help monitor different kinds of crops,  animals, and fishes. In the agrifood system, data can be captured from various sources, such as satellites, unmanned aerial vehicles (UAVs), and different sensors like the global positioning system (GPS) or cameras. Each of these data sources has its own properties that make them suitable for different applications. For example, data from satellites can help monitor growth, while images captured using UAVs are suitable for crop classification. It's essential to select the appropriate data sources for different agrifood applications. After determining the appropriate data source for a specific application, the second challenge is designing proper methods to efficiently exploit the data for prediction. The decision process of data and method selections based on specific applications has been shown as a mindmap on the project website due to the page limit. For instance, support vector machine (SVM) based methods~\cite{duan2019remote} are typically incorporated with RS data for rough agriculture classification. With the development of DL methods, specific models are leveraged for more classification tasks, such as recognizing different crops following the segmentation pipeline~\cite{yang2018high}. In addition, long short-term memory (LSTM) models are involved with spatial models to jointly process spatial and temporal data~\cite{zhang2017spatiotemporal}, such as soil or climate conditions that change over time, for more accurate prediction of crop yield. Moreover, as the rapid development of agriculture, animal husbandry, and fishery, the issues of data storage and processing pose additional challenges for allowing AI to transform agrifood systems.

In summary, the use of AI techniques in agrifood systems has brought significant benefits in terms of improving food security, reducing waste, and enhancing supply chain management. However, there are also challenges to be resolved, such as data collection, designing appropriate methods to leverage the data, and dealing with data storage and processing. Despite these challenges, recent progress in AI techniques, such as DL, has shown great potential in solving these issues and improving the performance of smart agrifood systems. Therefore, a comprehensive review of recent AI advancements in agrifood systems is essential to address research gaps, inspire further studies, and promote the application of advanced AI techniques, such as large-scale foundation models, to overcome challenges and unlock AI's full potential in this field. In Table~\ref{tab:abbreviation}, we summarize the abbreviations that will be used in this survey.

 \begin{table}[t]
 \caption{List of the abbreviations used in the main paper.}
 \resizebox{\linewidth}{!}{
\begin{tabular}{llll}
\hline
Abbreviation & Description  & Abbreviation & Description      \\ 
\hline
AdaBoost     & adaptive boosting                              &  LightGBM     & light gradient boosting machine                \\
AGB          & above-ground biomass                           &  LSTM         & ong short-term memory                          \\
AI           & artificial intelligence                        &  mAP          & mean average precision                      \\
AIoT         & artificial intelligence of things              &   Mask R-CNN   & mask region-based convolutional neural network \\
ANN          & artificial neural network                      &   mIOU         & mean intersection over union                   \\
BNN          & bayesian neural network                        &   ML           & machine learning                               \\
BPNN         & back-propagation neural network                &  MLP          & multilayer perceptron                          \\
MLR          & multiple linear regression                     &
CEC          & cation exchange capacity                       \\ MODIS        & moderate-resolution imaging spectroradiometer  &
CNN          & convolutional neural network                   \\  NDVI         & normalized difference vegetation index         &
DEM          & digital elevation model                        \\ NIR          & near-infrared                                  &
NN           & neural network                                 \\
DL           & deep learning                                  &  OLI          & operational land imager                        \\
DNN          & deep neural network                            &  OLS          & ordinary least squares                         \\
DT           & decision tree                                  &  PCA          & principal component analysis                   \\
EL           & ensemble learning                              &  PLSR         & partial least squares regression               \\
ELM          & extreme learning machine                       &  RF           & random forest                                  \\
EVI          & enhanced vegetation index                      &  RMSE         & root mean squared error                        \\
FCN          & fully convolutional network                    &  RS           & remote sensing                                 \\
GBDT         & gradient-boosting decision trees               & SAR          & synthetic aperture radar                       \\
GBRT         & gradient boosting regression tree              &  SOM          & soil organic matter                            \\
GEE          & google earth engine                            &  SURF         & speeded-up robust features                     \\
GLCM         & gray level co-occurrence matrix                &  SVM          & support vector machine                         \\
GLM          & generalized linear model                       &  SVR          & support vector regression                      \\
GPR          & gaussian process regression                    &  
GPS          & global positioning system                      \\
HDCUNet      & hybrid dilated convolution U-Net               &  TM           & thematic mapper                                \\
HOG          & histogram of oriented gradients                &  UAVs         & unmanned aerial vehicle                        \\
IoT          & internet of things                             &  VHR          & very high resolution                           \\
KNN          & k-nearest neighbor                             &  
LAI          & leaf area index                                \\ XGBoost      & extreme gradient boosting                      &
LASSO        & least absolute shrinkage selection operator    \\
\hline
\end{tabular}
}
\label{tab:abbreviation}
\end{table}

\subsection{Motivations of the Survey}

Although AI technology is greatly reshaping the landscape of agrifood systems and improving corresponding productivities, specific approach implementations and adopted data sources and processing ways in extensive subdivision applications that simultaneously involve agriculture, animal husbandry, and fishery, especially for intelligent interpretation methods adopting RS data, have not been systematically arranged and summarized, hindering related practitioners benefit from the engagement of advanced AI models. In addition, we also notice possible development prospects and potential challenges of advancing AI technology in the agricultural field. These reasons collectively prompt us to sort out relevant progress and complete this survey.

\subsection{Contributions of the Survey}

There are several excellent existing surveys regards using AI techniques in agrifood systems, a detailed discussion and comparison of which is provided below. Here, we specifically focus on applying the most recent AI techniques in agrifood systems from the data acquisition and method selection perceptive, respectively. To this end, about 200 studies are shortlisted and classified in this survey by using search keywords such as agrifood, RS, and AI. The main inclusion and exclusion strategies for the article are as follows: 1) The publication time of the article is limited from 2012 to the present; 2) Data obtained by RS techniques must be included in the research data; 3) AI methods must be used for research, containing traditional ML methods or DL methods that have become mainstream in recent years. Besides, an overview with discussions focusing on advanced research, potential challenges, and future research directions is also provided in this survey. The contributions of this survey can be summarized as follows.
1) We discuss how AI has transformed agrifood systems in various agrifood applications covering agriculture, animal husbandry, and fishery. 
2) We first review data acquisition in agrifood systems, including data sources, data storage, and data processing, and then review the recent progress in applying AI techniques in agrifood systems, covering a wide range of related topics.
3) We thoroughly discuss the potential challenges and future opportunities in applying AI in agrifood systems.

\subsection{Relationship to Related Surveys}
\subsubsection{Remote sensing}

Several existing surveys have examined the intersection of RS and agriculture \cite{survey_rs_agri_1,weiss2020remote,survey_rs_agri_3,survey_rs_agri_4}. For instance, Sishodia et al. \cite{survey_rs_agri_1} covers the history of RS and its impact on agriculture, as well as various vegetation indices that can be computed from RS images. Meanwhile, Khanal et al. \cite{survey_rs_agri_4} reviews and analyzes literature to identify recent developments in agriculture RS from temporal and geographical attributes. Other surveys have focused on specific RS data sources \cite{survey_rs_agri_8,survey_rs_agri_9,lu2020recent} or specific applications within agriculture \cite{survey_rs_agri_6,survey_rs_agri_7}. Additionally, some surveys have also reviewed specific AI models, such as CNNs, in the context of agriculture RS \cite{survey_rs_agri_5}. Overall, these surveys mainly provide detailed descriptions of the characteristics and applications of RS technology in agriculture. In contrast, this survey focuses on the employed extensive AI models in agrifood systems, where RS is regarded as one kind of important data source.

\begin{wrapfigure}[]{r}[0em]{0.45\textwidth}
    \centering
    \includegraphics[width=0.9\linewidth]{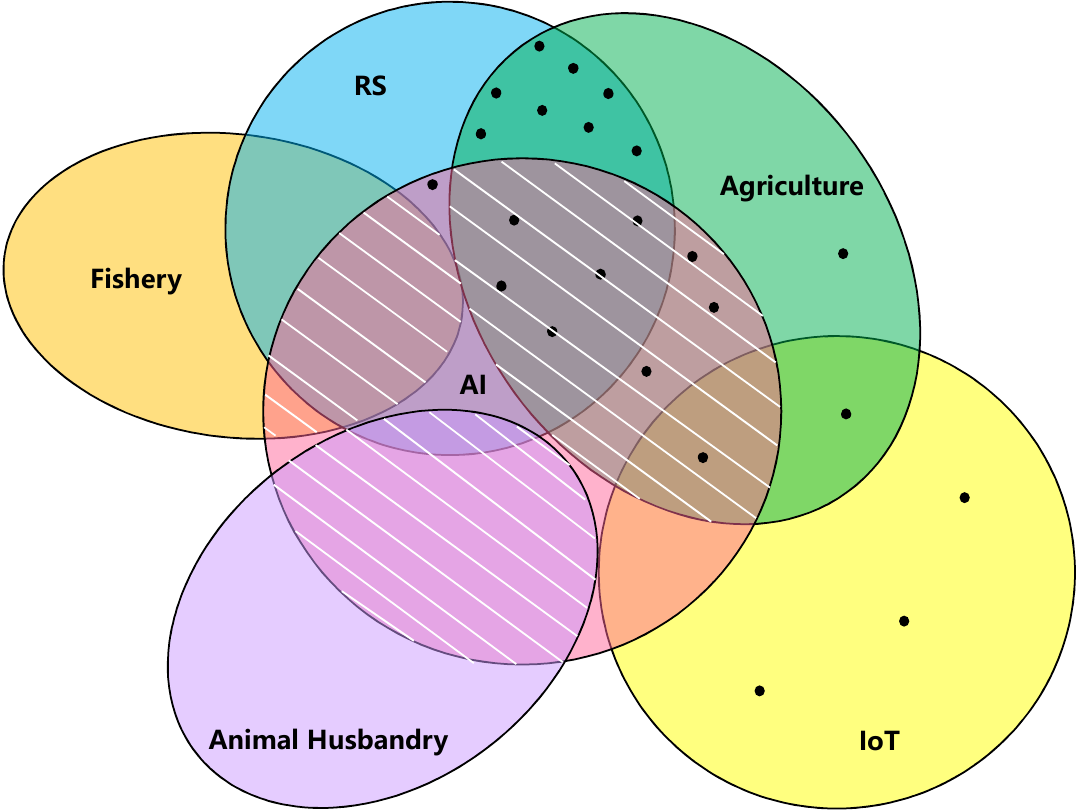} 
    \caption{The relationship between this survey and other surveys. The target domain of other surveys involved in Sec.~1.3 is represented as black points. The areas covered by white lines demonstrate the main theme of this survey.}
    \label{venns}
\end{wrapfigure}

\subsubsection{Internet of Things}

Internet of Things (IoT) represents an intelligent network that enables cyber-physical interactions by connecting numerous things with the capacity to perceive, compute, execute, and communicate with the internet. As presented in the recent surveys, IoT technologies have greatly advanced various areas \cite{stojkoska2017review,zanella2014internet,tokognon2017structural}, such as smart cities, smart homes, and smart health care. The smart agriculture with the help of IoT technologies has also been described in \cite{elijah2018overview}. Recently, Zhang et al. \cite{zhang2020empowering} discussed the possibilities that AI techniques can brought to the IoT area. Different from existing IoT-related surveys that mainly focus on the integration of IoT devices and different areas or technologies, this survey covers a wide range of topics of applying AI techniques in different agrifood fields, where IoT devices serve as the carrier for model and data.

\subsubsection{Agriculture}

As AI techniques continue to evolve within the field of agriculture, scholars have taken various perspectives in reviewing these methods. For example, Wang et al. \cite{wang2022review} has focused on the DL methods in multiscale (leaf-scale, canopy-scale, field-scale, etc.) agricultural crop sensing. They primarily review supervised learning, transfer learning, and few-shot learning methods. Kamilaris et al. \cite{kamilaris2018deep} conducts a comprehensive analysis of data resources and agriculture sectors for previous efforts employing DL methods. Zhai et al. \cite{zhai2020decision} provides an overview of decision support systems in agriculture, focusing on their systematic application in areas such as agricultural mission planning, water resources management, climate change adaptation, and food waste control. Compared to previous surveys that cover a broad range of topics in agriculture, some researchers have chosen to focus on specific fields within the agriculture. For example, Lu et al. \cite{lu2020survey} analyzes over 30 datasets for diverse agricultural topics, discussing the main characteristics and applications of each dataset, and highlighting key considerations for creating high-quality public agricultural datasets. Crop-related methods are reviewed in \cite{loey2020deep,zaji2022survey,hasan2021survey} while Koirala et al. \cite{koirala2019deep} reviews DL methods in fruit growing, recognition, and yield estimation. In summary, existing surveys focus on specific AI applications in agriculture, such as agricultural image datasets for computer vision and DL techniques for particular crops, while this survey provides a broader perspective. It covers recent advancements in both DL and traditional ML methods across various agricultural tasks and considers a wider agrifood system, encompassing animal husbandry and fishery.

For the convenience of readers' understanding, we clearly indicate the focus topic of this survey and the relationship to other surveys using a Venn diagram, as shown in Fig. \ref{venns}.

\subsection{Expected Audience of the Survey}

This survey can assist technical practitioners in related industries, such as engineers from agricultural intelligence companies and institutes, in finding appropriate AI techniques and data source types for subdivision applications. The experts of agricultural disciplines in universities and research institutions may also acquire useful insights into applying AI-based methods from this survey, for developing new agricultural production technologies and advancing the development of the whole agrifood systems. In addition, this survey can also facilitate the administrators in charge of agriculture departments in the government to focus on the outlooks and potential hidden risks of applying and generalizing AI technology, so as to make more suitable decisions. Finally, the media in the fields of agriculture, science, and technology can also benefit from this survey by learning about the advancement of the deep combination between AI and agriculture, and then popularizing the knowledge to the public including farmers, further promoting efficient deployments of AI techniques.

\begin{figure}[t]
    \centering
    \includegraphics[width=0.7\linewidth]{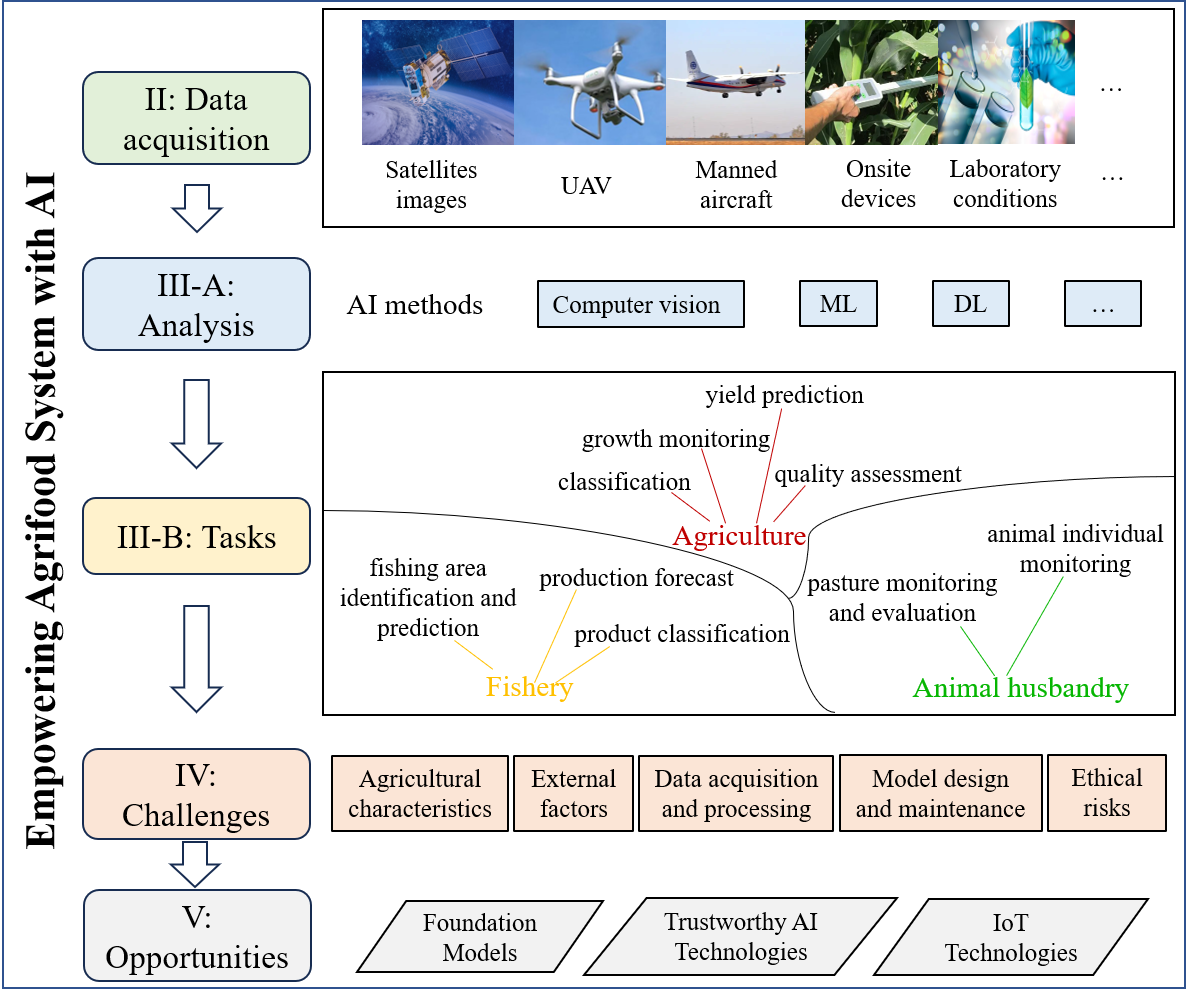} 
    \caption{Diagram of the section structure for this survey.}
    \label{section_corre}
\end{figure}

\subsection{Framework of the Survey}
AI methods in agrifood systems involve analyzing agricultural RS data from different platforms. This analysis uses computer vision, ML, data mining, and other advanced technologies to perform task-oriented research in specific agrifood systems, ultimately solving problems such as accurate prediction of crop yields, long-term monitoring of growth conditions, and precise classification of crop types. Fig.~\ref{section_corre} shows the general framework for this survey. Sec.~2 introduces the collection of different types of agricultural data from various platforms, including space platforms (e.g., satellites), aerial platforms (e.g., UAVs), ground platforms (e.g., measuring instruments), and laboratories. In addition, other specific datasets are available. Based on these agricultural RS data, in Sec.~3, we review different AI technologies, such as ML and DL in computer vision for agrifood data analysis. Aiming at different sub-tasks in the agrifood system, we investigate many approaches in corresponding fields and analyze the effect of related methods. Then, we point out the challenges that still need to be addressed in Sec.~4, such as the characteristics of agriculture, the influence of external factors, and moral hazard. Sec.~5 is a summary of some potential research directions that currently exist, including foundation models, trustworthy AI, and IoT technologies. Finally, Sec.~6 concludes the survey.

\section{Data Acquisition in Agrifood Systems}

In this section, we first provide a detailed introduction to the types and characteristics of data adopted in existing agrifood systems. Then, common storage and processing approaches for these data before utilization are presented.

\subsection{Data Sources}
\subsubsection{Satellites}

Satellite RS technology has become a widely used tool for agrifood monitoring due to its vast observation range and timely data updates. Satellite images provide essential multi-source, multi-temporal, and multi-resolution data for monitoring agricultural production, animal husbandry, and fishery. RS satellites are typically classified as active or passive based on their onboard sensor data acquisition methods. Passive ones usually refer to optical RS satellites, which offer rich spectral and textural features but can be severely impacted by cloudy, rainy, and foggy weather conditions. While active RS technology mainly involves the synthetic aperture radar (SAR), which can operate all around the clock, detecting land surface information by receiving reflected echoes from actively emitted microwaves.

\par SAR distinguishes different features by analyzing their backscattering characteristics. Because the emitted microwaves are sensitive to moisture, SAR is often used to retrieve soil and vegetation moisture levels \cite{tripathi2022deep}. When combined with optical images, it allows for more accurate and refined crop classification \cite{adrian2021sentinel,sun2019using} or monitoring \cite{chen2021estimating,bahrami2021deep}. Additionally, SAR can identify surface deformation information, making it useful for crop canopy monitoring. Currently, the most popular SAR satellite is the Sentinel-1, which is favored due to its free accessibility.

\begin{table}[b]
\centering
\caption{Details of existing optical remote sensing satellites.}
\resizebox{0.8\linewidth}{!}{
\begin{tabular}{ccccc} 
\hline
 Satellite/Sensor &  Resolution(m)  &  Number of bands  &   Wavelength range(nm)     &  Reference \\ \hline
 Sentinel-3       &  1200      &  21   &  400-1020     &  \cite{guzinski2019evaluating} \\
 MODIS  &  250/500/1000      &  36   &   400-14400     &  \cite{hao2022estimation,guzinski2019evaluating,jeong2022predicting,zhou2022integrating,sun2020multilevel,johnson2016crop,ali2016modeling} \\
 Sentinel-2       &  10/20/60     &13  &  400-2400     &   \cite{zhou2022research,lin2020continuous,kayad2019monitoring}   \\ 
 HJ-1A*, HJ-1B      &  30     &  115    &  450-950          &  \cite{jiang2018method,zhou2016estimation} \\ 
 EO-1*             &  30     &  242   &  356-2577   &  \cite{bhosle2019evaluation} \\
 Landsat 4-5  &  30       &  7   &  450-2350   &  \cite{samui2012statistical} \\
 Landsat 7  &  15     &   8    &  450-900   &  \cite{chen2022remote,shuai2022subfield} \\
 Landsat 8  &  15     &   11    &  430-12510  &  \cite{shuai2022subfield,li2020adversarial,xu2020deepcropmapping,zhang2019prediction} \\
 ASTER Terra  &  15    &   14    &   520-11650 &  \cite{pena2014object} \\
 GF-1             &  2 (PAN), 8 (PMS)    &  5  &  450-900 &  \cite{liang2021semi} \\
 GF-2             &  1 (PAN), 4 (PMS)  &   5   &  450-900 &  \cite{cheng2020research} \\
 GF-6             &  2 (PAN), 8 (PMS)  &   5   &  450-900 &  \cite{sun2021mapping}\\
 RapidEye         &  5     &  5   &  400-850          &  \cite{bahrami2021deep} \\
 PlanetScope      &  3     & 4    &   455-860         & \cite{parente2019next} \\
SPOT-5           & 2.5     &5    &   490-1750          & \cite{dos2013interactive} \\
 QuickBird  & 0.61   &  8  &  435-11190 & \cite{dos2013interactive} \\
world-view3 & 0.31  &   29   &  405-2365   & \cite{adrian2021sentinel,dela2021remote} \\
AVIRIS*(airborne)  & 20   &  224   &  200-2400  & \cite{bhosle2019evaluation,2021FullyContNet,moreno2014extreme} \\
\hline
\multicolumn{5}{l}{* denotes containing hyperspectral sensor; PAN: Panchromatic; PMS: Panchromatic and Multispectral}
\label{satellites}
\end{tabular}
}
\end{table}

\par
\begin{figure}[t]
    \begin{minipage}[t]{0.48\textwidth}
        \centering
        \includegraphics[width=\linewidth]{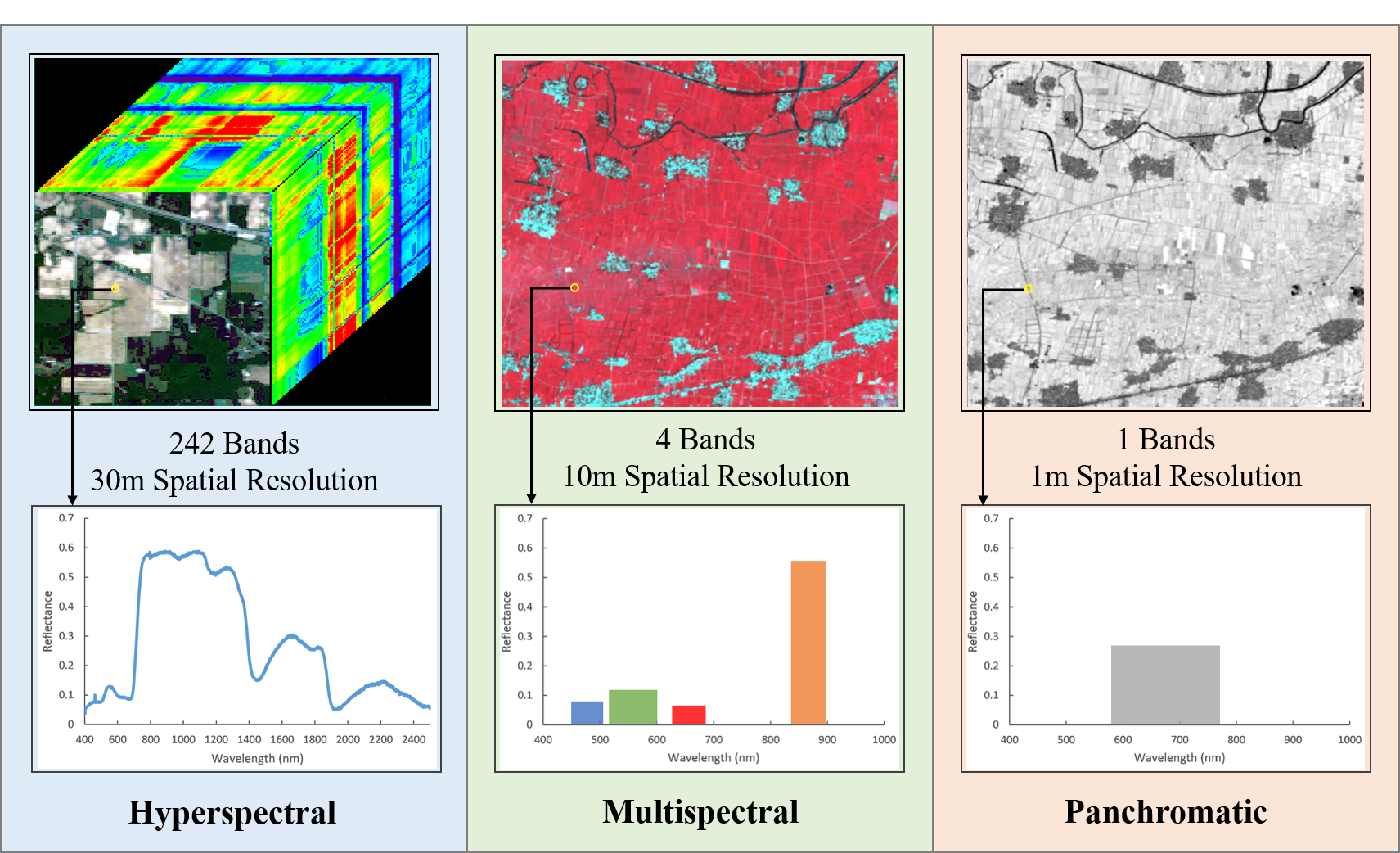}
        \caption{Example of optical images: Hyperspectral, Multi-spectral, and Panchromatic.}
    \label{YY_satellites}
    \end{minipage}    
    \hfill
    \begin{minipage}[t]{0.48\textwidth}
        \centering
       \includegraphics[width=\linewidth]{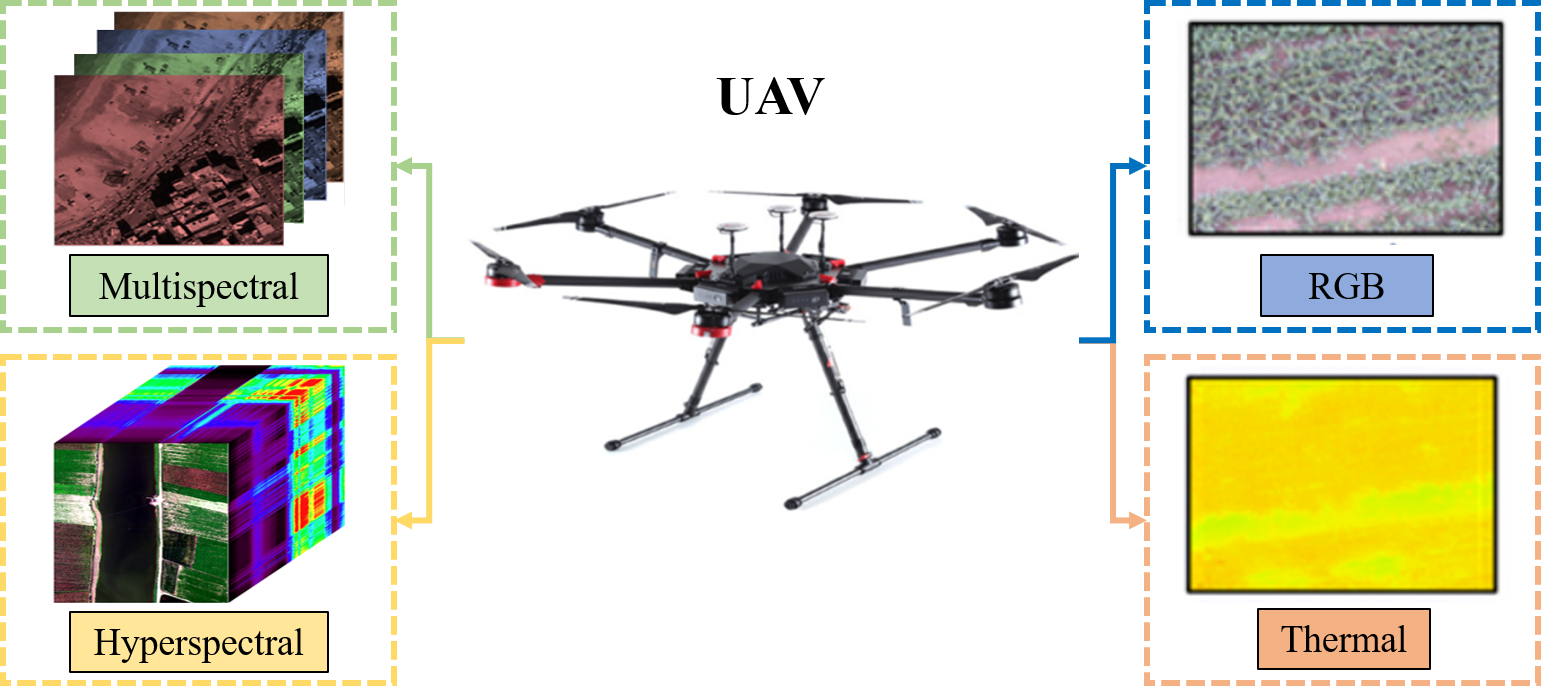}
       \caption{UAVs equipped with multiple sensors simultaneously.} 
      \label{UAV} 
    \end{minipage} 
\end{figure}

 Images obtained by optical sensors can mainly be divided into three types: multispectral, hyper-spectral, and panchromatic, as shown in Fig.~\ref{YY_satellites}. Hyperspectral sensors receive abundant spectral information, allowing the extraction of a continuous spectral curve of a broad wavelength range for each pixel. Thus, the obtained images are appropriate for distinguishing vegetation with similar visual characteristics. This is very useful in fine-grained vegetation classification \cite{bhosle2019evaluation,2021FullyContNet,moreno2014extreme}. Multispectral images are most commonly used for agricultural classification and yield prediction. Moderate-resolution imaging spectroradiometer (MODIS), Landsat-8, and Sentinel-2 are the three most popular medium-resolution multispectral satellites for agrifood monitoring \cite{hao2022estimation,lin2020continuous,shuai2022subfield} since their data are easy to acquire and rich in vegetation-sensitive bands.

\par Some very high resolution (VHR) satellites such as SPOT5 \cite{dos2013interactive}, QuickBird \cite{dos2013interactive}, and RapidEye \cite{bahrami2021deep} are suitable for fine-grained classification and field-level monitoring. However, due to their commercial availability and high cost, these satellites have relatively fewer applications in agrifood monitoring. Table~\ref{satellites} provides details of common optical RS satellites and their resolutions.

\subsubsection{Unmanned Aerial Vehicles}
UAV platforms can provide ultra-high spatial resolution images for agrifood systems \cite{varela2018early}. Compared to traditional airborne and satellite platforms, UAV images offer finer spatial, spectral, and temporal resolutions, making them more suitable for precision agriculture. UAV platforms offer great flexibility in terms of acquisition operation, flight cycle, flight altitude, and geographic coverage \cite{lu2020recent}. They are also able to significantly reduce the impact of atmospheric disturbances on the generated images and eliminate cloud cover issues that commonly affect optical satellites and high-altitude airborne optical sensors. UAVs can be compatible with multiple types of sensors, as shown in Fig. \ref{UAV}, allowing for lower costs and obtaining diverse monitoring data. Therefore, many studies simultaneously utilize UAV multi-source data \cite{pang2020improved,oliveira2020machine,zhu2019estimating,maimaitijiang2020soybean}. For instance, Pang et al. \cite{pang2020improved} combined RGB and multispectral images to improve early-season maize strain calculations, while Oliveira et al. \cite{oliveira2020machine} jointly utilized RGB and hyperspectral imagery to assess the quantity and quality of pasture. Maimaitijiang et al. \cite{maimaitijiang2020soybean} evaluated the ability of UAV-based multimodal data fusion (incorporating RGB, multispectral, and thermal sensors) to estimate soybean grain yield. Zhu et al. \cite{zhu2019estimating} utilized image data acquired by multispectral and LiDAR sensors as source data for maize above-ground biomass (AGB) estimation.

\subsubsection{Manned aircraft}

\par  Besides UAV, aerial imagery can also be obtained from sensors mounted on a piloted (manned) aircraft. Mattupalli et al. \cite{rs10060917} tested two images obtained by different sensors that are mounted on UAVs and manned aircraft, respectively. The images are then used to classify three crop types by ML algorithms. The imagery from the manned aircraft achieves better overall accuracies (OAs).

\subsubsection{Onsite devices}

\par
\begin{figure}[t]
    \centering
    \includegraphics[width=0.8\linewidth]{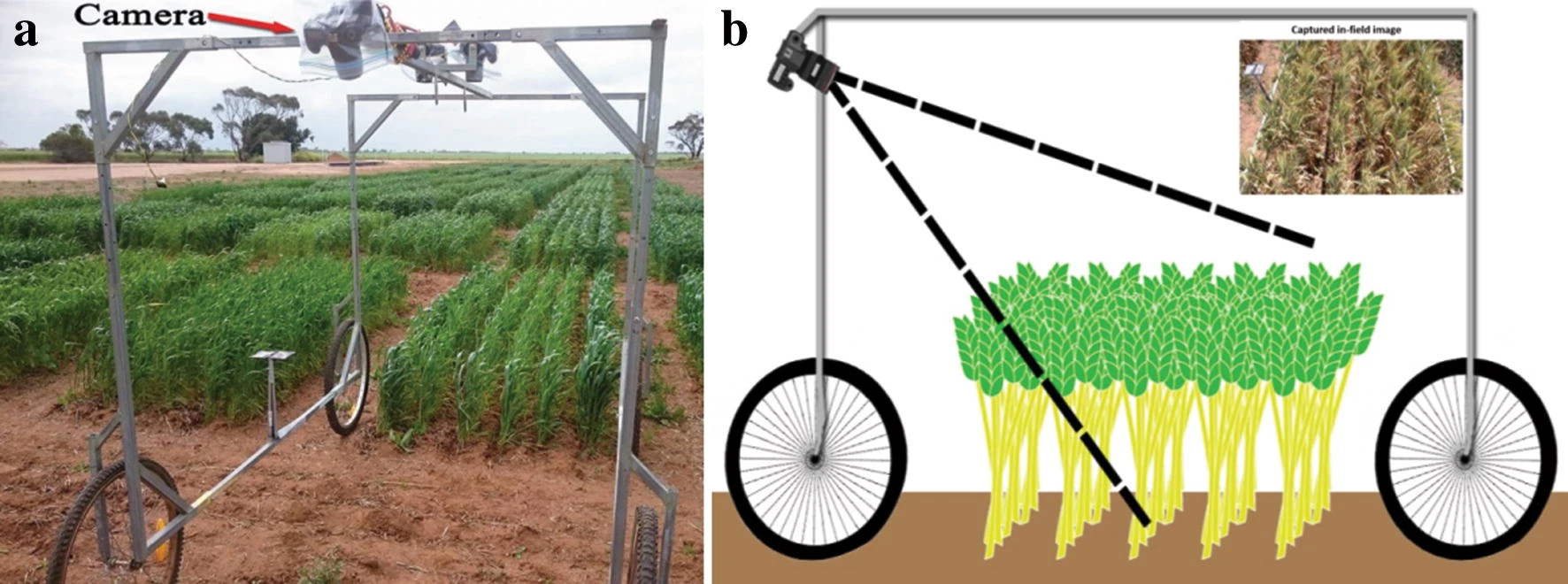} 
    \caption{Visualization of onsite devices' data acquisition process~\cite{hasan2018detection}. (a) Field photo of the onsite device tool. (b) The schematic of the tool and a sample image taken with the oblique-view camera.}
    \label{on_site}
\end{figure}

 Onsite devices such as ground height spectrometers, handheld spectrometers, handheld GPS, multispectral cameras, thermal imaging cameras as well as digital cameras have been used for various specific-purpose agricultural studies. For example, Hasan et al. \cite{hasan2018detection} uses a high-resolution RGB camera mounted on a land-based imaging platform to acquire images for one season to estimate wheat yields. Fig.~\ref{on_site} presents field photographs of the data acquisition process as described in \cite{hasan2018detection}. It is comprised of a steel frame and four wheels with a central overhead rail for mounting imaging sensors. Hu et al. \cite{hu2019pixel} manually takes high-resolution images of wheat above the canopy from budding to flowering for breeding selection. 

\subsubsection{Other sensors}

Some specialized devices, such as the time domain reflectometer (TDR) for monitoring soil moisture dynamics of the surface and root zone \cite{babaeian2021estimation} and the RSX-1 gamma radiation detector for measuring the amount of radiation in the soil \cite{al2021impact}, are also employed in agrifood systems. Fig.~\ref{ll_TDR} presents the field photo of monitoring soil moisture with TDR. The observed data usually has high precision and is always adopted as the ground truth values for training and improving AI models. However, when applied to large-scale research, it will face the defects of high cost and low efficiency. At the same time, the point-like results obtained from such sensors cannot effectively cover all areas.

\subsubsection{Laboratory conditions}

\begin{figure}[!tbp]
    \begin{minipage}[t]{0.35\textwidth}
        \centering
        \includegraphics[width=0.93\linewidth]{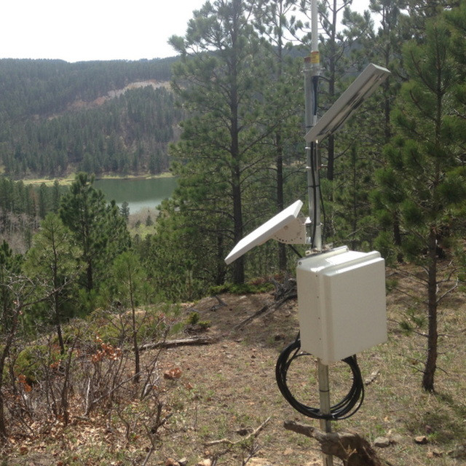}
        \caption{Field photo of TDR equipment monitoring soil moisture \cite{TDR2024}.}
    \label{ll_TDR}
    \end{minipage}    
    \begin{minipage}[t]{0.5\textwidth}
        \centering
       \includegraphics[width=0.98\linewidth]{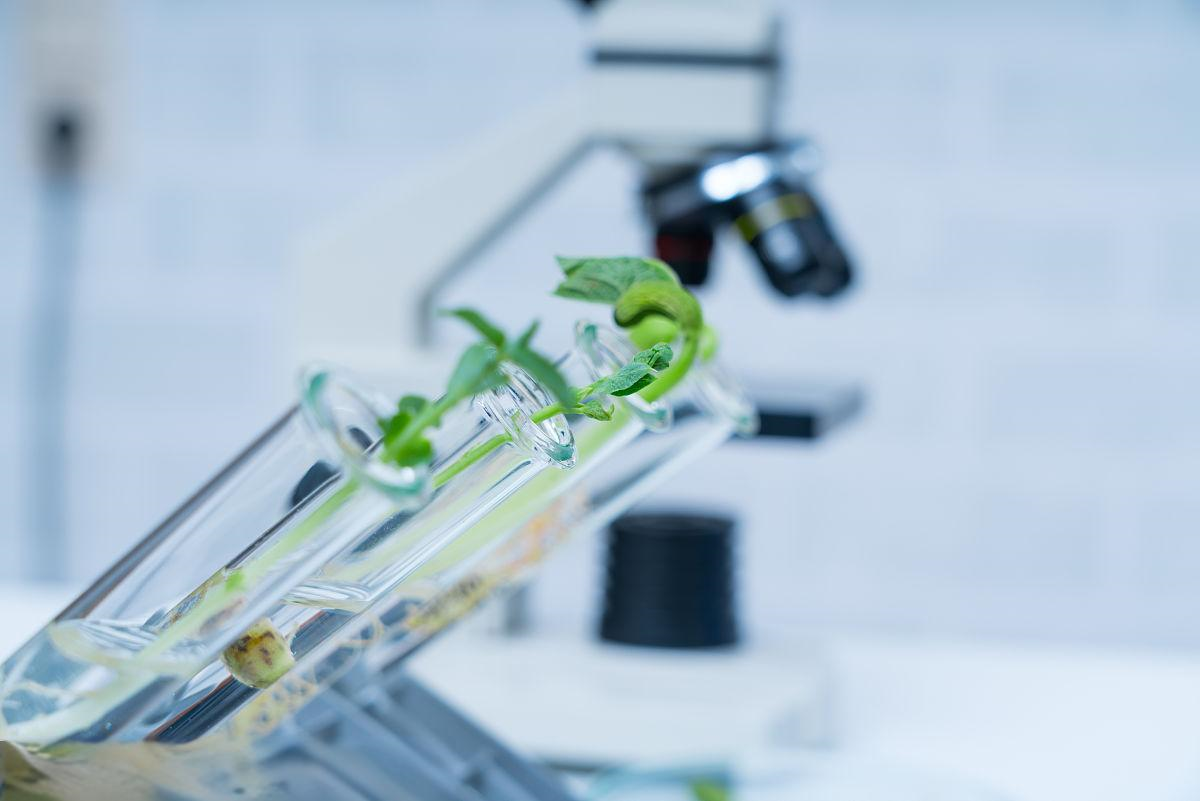}
       \caption{Monitoring plant growth status in the laboratory \cite{Seed2024}.}
      \label{ll_laboratory} 
    \end{minipage} 
\end{figure}

Due to various factors such as field environment, equipment, experimental operation, and data analysis, much experimental data cannot be obtained directly in the field or through RS and other approaches. At this time, representative research data can be obtained by analyzing collected field samples in the laboratory. For example, in the laboratory, Tripathi et al. \cite{tripathi2022deep}, Babaeian et al. \cite{babaeian2021estimation} and Khanal et al.\cite{khanal2018integration} perform soil organic matter content measurements, Li et al. \cite{li2022improving} uses the Kjeldahl method to measure the Nitrogen (N) content for wheat leaves, while Fu et al. \cite{fu2020wheat}, Zhou et al. \cite{zhou2016estimation} and Zhang et al.\cite{zhang2018capability} obtain the dry weight of wheat by desiccation. Fig.~\ref{ll_laboratory} shows the monitoring of plant growth status in the laboratory.

\subsection{Data Storage and Processing}
\subsubsection{Data storage}

Satellites are typically operated by governments and the observed RS images are stored in ground station data centers, which can be accessed by visiting the appropriate websites, such as the U.S. Geological Survey and European Space Agency. Additionally, large-scale geospatial data analysis platforms like google earth engine (GEE) also offer preprocessed images for users to access. Recently, some commercial companies have started manufacturing satellites, particularly for VHR applications, and they store the resulting data in private databases \cite{datasource_agri_1, datasource_agri_3}. UAVs, onsite devices, and other sensors can collect data at any time and in any location, which are cached directly on the onboard disk or sent to a remote data center. In laboratory settings, data are typically recorded manually or by using specialized instruments \cite{datasource_agri_2, lu2020survey}.

\subsubsection{Data processing}
Proper preprocessing of data is essential before utilizing it, especially when it comes to optical satellite images. Radiometric calibration and correction, atmospheric correction, and geometric correction are the most crucial steps involved in this stage. Initially, digital numbers of original images are transformed to surface reflectances, and the influences of atmosphere and earth curvature during observations are removed. Clouds, shadows, water, and snow may also be eliminated if necessary. Ortho rectification is conducted to correct the influence of topography, and image mosaicking is required if the study area spans different tiles. Band selection and clipping are usually implemented, and sometimes, the geographic coordinate system needs to be converted. Resampling, multi-temporal interpolation, image fusion, and different index calculations are conducted for specific applications in agrifood systems. Preprocessing SAR images requires additional steps compared to optical data, as speckle noise caused by systematic errors must be removed through multi-looking correction or filtering. For UAV data, standard photogrammetric procedures are used to preprocess RGB images, which involves aligning and merging images and then producing a series of products, such as the digital elevation model with the help of ground control points. When dealing with multispectral data, radiometric and lens corrections are also necessary. Preprocessing hyperspectral data requires further treatments to address sensor, atmosphere, and solar light issues. Preprocessing data obtained from handheld cameras is similar to UAV images, while data from other onsite devices will require specific processing depending on the circumstances. For instance, spectral data obtained from a spectrometer typically requires denoising and smoothing. In some cases, such as data obtained in laboratory conditions, preprocessing may not be necessary since the data is measured as a reference of the predicted value in AI models. Most of the data preprocessing approaches have been integrated into open-source or commercial software, and readers can refer to public collections\footnote{https://github.com/satellite-image-deep-learning} for details. In addition, regarding the preprocessing methods that should be taken for specific data, we visualize corresponding relation mapping of the types and preprocessing for data used in existing agrifood systems, as shown on our project website.

\section{Progress Review of AI Methods in Agrifood Systems}
This section will provide an overview of AI methods used in agrifood systems. Firstly, we will briefly categorize existing AI methods into two groups, i.e., traditional ML and DL. Secondly, we will conduct a comprehensive literature review on the application of AI methods in various agricultural sectors such as crops, animal husbandry, and fisheries. Given that the same AI techniques can be used for diverse applications in agriculture, we will organize the review according to an application taxonomy that encompasses both traditional ML and DL approaches for various applications.

\subsection{Brief Categorization of Existing Methods}

\subsubsection{Traditional machine learning methods}

There are a large number of long-standing traditional ML methods popular, such as k-nearest neighbor (KNN), Decision Tree (DT), gaussian process regression (GPR), SVM, RF, extreme gradient boosting (XGBoost), etc. In the field of agriculture, traditional methods primarily focus on multiple factor classification tasks, and a series of regression analyses represented by metric prediction and estimation. For example, DTs \cite{moreno2014extreme}, SVMs \cite{chen2021estimating}, and gradient-boosting decision trees (GBDT) \cite{lozano2022remote} have been employed by researchers to identify a range of crops, including coffee, wheat, corn, rice, etc, based on various related factors. Besides,  linear regression (LR) \cite{fu2020wheat}, ordinary least squares(OLS) \cite{zhu2019estimating}, and KNN \cite{cheng2022estimation} are common techniques used to analyze various types of images for prediction and estimation tasks such as soil moisture monitoring, yield prediction, and biomass estimation. Apart from crops, traditional ML methods also play important roles in various topics in animal husbandry including pasture monitoring and sickness detection. For fishery, they are improving fishing area identification and production forecast domains.

\subsubsection{Deep learning methods}
\begin{wrapfigure}[]{r}[0em]{0.45\textwidth}
    \begin{minipage}{1\linewidth}
        \centering
        \includegraphics[width=\linewidth]{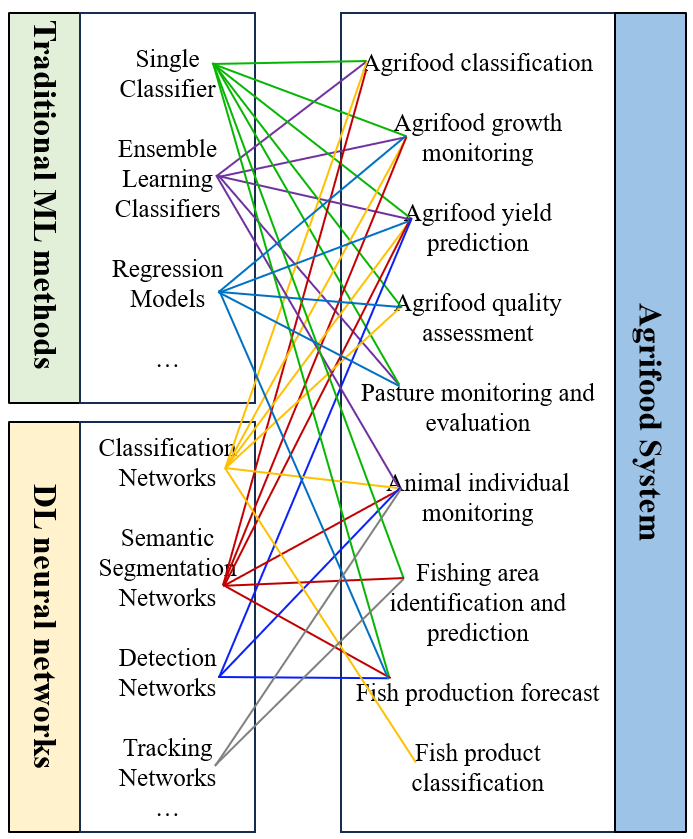}
    \end{minipage}    
    \caption{Correlation between AI methods and applications in the agrifood system.}
    \label{yy_correlation}
\end{wrapfigure}

With the continuous increase of computing resources and data scales, DL has emerged as a powerful tool for deep feature extraction from raw images, surpassing traditional ML approaches in terms of performance. Researchers thus have increasingly focused on exploring the use of deep neural networks in various applications in agriculture, resulting in numerous methods for tackling diverse tasks. In the context of computer vision, deep neural networks typically comprise a backbone and task-specific heads that specialize in different tasks. In agricultural applications that involve structural data, such as images, the most commonly used backbones are CNNs \cite{jeong2022predicting}, 3D CNNs \cite{nevavuori2020crop}, and LSTM \cite{zhang2019combining} networks. CNNs are good at handling tasks with 2D images such as multi-crop classification, while the latter two models are popular in tasks involving temporal data, such as predicting crop yields or tracking locations for moving animals. Apart from them, the recent emergence of Vision Transformers (ViT)  has attracted significant attention among researchers in many fields like crop classification and weed extraction \cite{reedha2022transformer}. These powerful models have been employed for a range of agricultural tasks, where researchers often utilize different task heads to generate the required output based on the specific problem formulation. Accordingly, these models can be categorized from a fundamental task perspective. For example, classification models aim to predict the class of given images, while segmentation models seek to predict the labels of each pixel. Detection models need to generate the objects' locations as well as the corresponding classes while tracking models estimate the trajectory of an arbitrary target object. These models are capable of formulating tasks related to crops, animals, and fishery, and have demonstrated effectiveness in addressing a range of problems in these domains. Fig.~\ref{yy_correlation} briefly shows the relationship between AI approaches in agrifood systems and the corresponding applications.

\subsubsection{Model evaluation}
Compared to traditional methods, AI methods have achieved great progress in terms of accuracy, scalability, and cost. Taking the agrifood classification task as an example, metrics such as overall accuracy (OA) and Kappa coefficient (Kappa) are usually used to evaluate the classification performance \cite{moreno2014extreme, liao2020synergistic, lozano2022remote}. For instance, the AI models achieved over 90\% OA and Kappa values on the WHU-Hi dataset \cite{zhong2020whu}. In addition, AI methods represented by large-scale foundation models can better adapt to new scenes over traditional approaches by constantly pre-training with new data. In terms of cost, traditional methods require substantial manual efforts for data annotation, while AI methods can leverage unlabeled data for training.

\subsection{Artificial Intelligence Methods in Agriculture}

\subsubsection{Agrifood classification}
\par RS technologies have been widely applied to agriculture, since they provide multi-source, multi-temporal, and multi-resolution data for large-scale, all-weather automated agrifood monitoring. According to the research purposes and scales, RS-based agrifood classification mainly includes 1) cropland recognition, which aims at identifying all croplands in study areas, regardless of the specific crop type; 2) multi-type classification, which aims to classify the specific crop type; 3) single-type identification, which aims at identifying only interested crop type. Please refer to the appendix for distinguishing different classification types through visualizations. Cropland recognition is suitable for large-scale provincial-level or county-level tasks. Multi-type classification and single-type identification provide more detailed cropland information and are usually used for field-level or provincial-level tasks. Generally, the methods for these tasks consist of feature extraction and classification processes. Recently, an increasing number of ML and DL methods have been proposed for agrifood classification and delivered good performances.

\par In the early years, most of the methods utilized band algebra and band transformation (e.g., difference vegetation index, enhanced vegetation index (EVI), normalized difference vegetation index (NDVI) \cite{zhang2017spatiotemporal,pena2014object,chang2007corn}) to extract features for classification. However, these methods often result in noisy and inaccurate classification due to manual threshold setting and subjective expert experience. In order to utilize the abundant inter-classes, sample-wise and temporal relationships introduced by RS images, traditional ML methods \cite{sharma2022above} have been applied in agrifood classification as they can explore the latent and complex information between samples belonging to different categories. At first, researchers usually use a single ML classifier for classification. Moreno et al. \cite{moreno2014extreme} uses C4.5 DT for hyper-spectral soy-bean mapping. Although the rules generated by C4.5 DT are easy to understand, the calculation has low efficiency when coping with large datasets. Kussul et al. \cite{kussul2017deep} adopts multilayer perceptron (MLP) for multi-temporal crop classification, but its performance is limited by the small number of latent layers, which may cause the model cannot learn highly representative and discriminative features.

\par Most of the traditional machine learning methods can capture complex nonlinear features in data, but a single method, like LR \cite{fu2020wheat}, DT \cite{moreno2014extreme}, may be difficult to fit or fall into the local optimum when dealing with large-scale datasets, resulting in a high recognition error. Therefore, researchers try to integrate multiple classifiers to comprehensively utilize the advantages of different classifiers \cite{pena2014object, alsuwaidi2018feature}. The idea of integrating multiple classifiers has later extended to the field of ensemble learning (EL), which is nowadays one of the most representative and effective methods. Typical EL methods like RF \cite{van2018synergistic}, and GBDT \cite{lozano2022remote} are widely applied in agriculture classification and outperform other single classifier-based methods. In addition, researchers are continuously working on constructing strong and robust feature extractors for classification. Recently, DL methods have gained increasing attention in crop classification and identification tasks, e.g., multi-type crop classification \cite{zhao2019evaluation,lei2021docc}, sorghum identification \cite{wang2019mapping}, soybean and maize crop \cite{xu2020deepcropmapping} mapping and so on. Most of the neural networks applied in agriculture classification are semantic segmentation networks, as crop classification and identification tasks need pixel-level labels. Huang et al. \cite{huang2021depth} verifies the effectiveness of four semantic segmentation models on identifying tobacco planting areas through UAV images. State-of-the-art networks such as ViT \cite{reedha2022transformer} and TransUNet \cite{niu2022hsi} are also applied in multi-type crop classification.

\par Generally, RS agrifood classification based on ML and DL methods can handle input images in various resolutions, so it can satisfy diverse research scales such as field-level, provincial-level, and even country-level and meet the requirements of different classification tasks. Nevertheless, there is still a large room for advancing RS agrifood classification, by leveraging more powerful neural networks and learning methods.

\subsubsection{Agrifood growth monitoring}
\par
\begin{figure*}[!t]
    \begin{minipage}{1\linewidth}
        \centering
        \includegraphics[width=0.8\linewidth]{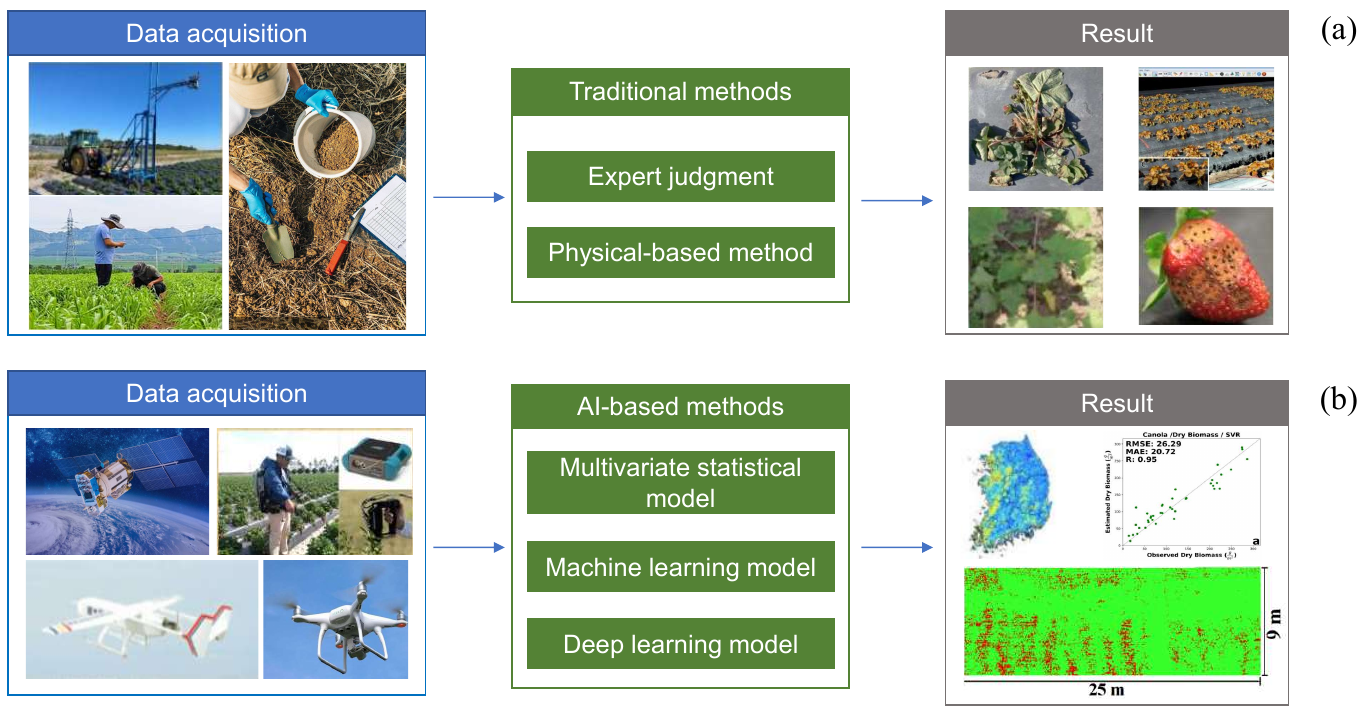}
    \end{minipage}    
    \caption{Comparison of (a) traditional methods and (b) AI-based methods in Agrifood growth monitoring.}
    \label{ll_traditional and AI technology}
\end{figure*}

 Crop growth state is a crucial component of agricultural data, as it provides insights into the yield of the crop. Crop growth is affected by many factors, such as light, temperature, weather, soil, water, gas (CO2), fertilizer, diseases, insect pests, and so on. In the early stages of crop growth, it is mainly reflected by the quality of crop seedlings, while in the middle and late stages of crop growth and development, it mainly reflects the development of crop plants and the specific characteristics of high and low yields. Although the growth of crops is an extremely complex physiological and ecological process that involves many factors, it still can be characterized by some metrics that can reflect its growth state or are closely related to its growth characteristics, such as nitrogen (N) \cite{li2022improving}, phosphorus (P) \cite{gao2019modeling} and potassium (K) \cite{misbah2021multi}. In the past, relevant crop growth state factors and environmental attribute factors were mainly monitored and collected manually, and then comprehensively analyzed by experts according to their experiences or using statistical models containing multiple variables. The obtained results may be subjective and inaccurate, which could not meet the requirements of intelligent and precise agriculture (see Fig.~\ref{ll_traditional and AI technology}).

\par As an important subset of AI, ML techniques have proven to be a promising solution for addressing the spatial analysis of big data and solving nonlinear problems, especially when the extent of theoretical knowledge of a problem is incomplete \cite{lary2016machine} or when statistical presuppositions are unreliable or not known \cite{dou2019assessment}. At present, ML technologies have made considerable achievements in the field of crop growth monitoring. We have summarized some typical examples of agrifood growth monitoring based on RS imagery and statistical ML algorithms in the appendix. In the following text, the applications of AI methods on different sub-directions of agrifood growth monitoring will be sequentially described.

\par \textbf{Environmental monitoring:} The environment of crop growth encompasses the external natural conditions of the living space of crops, including the effects of various natural environmental conditions and other biological organisms. Among these factors, soil moisture and the contents of soil organic matter (SOM), cation exchange capacity (CEC), Mg, K, and pH are key factors that significantly impact crop growth.

\par Monitoring soil moisture using RS information is crucial for precision agriculture management, particularly in regions with limited water resources. Compared to ground measurement methods, RS-based monitoring provides a more cost-effective means and has been increasingly used in large-scale soil moisture monitoring tasks \cite{shin2013development, sadeghi2017optical, babaeian2019ground}. Currently, RS-based soil moisture monitoring methods mainly rely on physical-based models, such as radiative transfer models, or statistical ML models. Physical-based models establish the functional relationship between soil moisture content and various spectral indices obtained from RS. These models require accurate parameters such as soil characteristics, land cover characteristics, vegetation biochemistry, and biophysics. Common methods include thermal inertia, crop surface temperature, gravity soil moisture measurement, soil water balance calculation \cite{ihuoma2017recent}, and so on. Although physical-based models can provide accurate information for soil moisture, obtaining parameter quantities makes it difficult to monitor the spatial and temporal distribution of soil moisture over vast areas. Additionally, collecting soil moisture data and laboratory measurements through sampling points is expensive, time-consuming, and labor-intensive \cite{shin2013development}, which are unsuitable to apply in a large range area.

\par Compared to physical models, statistical ML models learn the complex interactions between the plant-soil-atmosphere continuum by establishing the relationship between spectral signals and plant characteristics, and then fitting the soil moisture in the study area. Predictors such as canopy reflectance can be directly associated with soil moisture as the target variable through regression models. For instance, Babaeian et al. \cite{babaeian2021estimation} jointly utilize UAV data and physical and hydraulic soil information to estimate soil moisture in the root zone. They adopt various algorithms such as artificial neural network (ANN), generalized linear model (GLM), gradient elevator, RF, and the EL methods that are included in an automatic machine learning (AutoML) platform. Cheng et al. \cite{cheng2022estimation} employs multi-modal data fusion and four ML algorithms (partial least squares regression (PLSR), KNN, RF, and back-propagation neural network (BPNN)) to estimate field soil water content. Apart from using satellite images and UAV images, SAR images have also been widely used to estimate soil moisture in crop areas. Chen et al. \cite{chen2021estimating} collects quad polarization's RADARSAT-2 data and 240 sample plots in the study area to compare the performance of three advanced ML models (namely, SVM, RF, and gradient boosting regression tree (GBRT)) in estimating soil moisture. Their results demonstrate that combining polarization decomposition parameters with ML and feature selection methods can effectively estimate soil moisture with high accuracy. This method is useful in monitoring soil moisture across the entire farmland during the growing season.

\par Other than soil water content, various soil chemical characteristics such as SOM, CEC, Mg, K, and pH also play a crucial role in affecting crop yield. In a study of \cite{khanal2018integration}, linear regression and five ML models, namely, RF, neural network (NN), SVM, gradient lifting model, and Cubist are used to predict the soil properties of seven farmlands near Ohio. While Zhang et al. \cite{zhang2019prediction} uses NDVI time series data and stepwise linear regression, PLSR, SVM, and ANN models to map soil organic carbon in Honghu City, where the ANN demonstrates the best performance.

\par \textbf{Crop growth state monitoring:}  N, P, K, and other essential elements are crucial for crop growth and play a vital role in plant nutrition. These elements often affect crop yield due to insufficient supply in the soil. However, traditional crop nutrition diagnosis relies on destructive on-site sampling and slow laboratory analysis \cite{vigneau2011potential}. In addition, the nutrient concentration measured at each point may differ significantly from the actual nutrient content in the farmland \cite{berger2020crop}. What's more, variations in crop nutrient content can lead to changes in crop morphology and leaf color, suggesting that spectral information of crops can be variable \cite{haboudane2002integrated}. These findings provide a theoretical basis for monitoring crop growth using RS technology, as it allows for non-destructive and continuous monitoring of crop nutrient status.

\par Commonly used crop monitoring and simulation methods include radiation transfer, process-based, and empirical statistical models. Jay et al. \cite{jay2017retrieving} notes that radiation transfer and process-based crop models are considered universal because they are developed based on physical laws, and ecological and physiological principles. However, obtaining high-resolution inputs and parameters for these models can be challenging \cite{duan2014inversion}. In contrast, Cai et al. \cite{cai2019integrating} highlights that empirical methods directly link crop variables with RS indicators, which is particularly useful for growth processes with limited information about potential mechanisms. Moreover, empirical study \cite{weiss2020remote} has shown promise that the ML/DL methods based on AI technology are proved to have similar simulation abilities compared to radiation transfer models in properly simulating reality. Later, Wang et al. \cite{wang2021estimation} evaluates the nutrient elements in rice crops at different growth stages using univariate regression models, multivariate calibration methods, PLSR, and several ML methods, including ANN, RF, and SVM, based on UAV hyperspectral images. Their experiments have demonstrated that PLSR models and ML methods can yield improved outcomes in circumstances where multiple growth stages are involved or there is a lack of phenological information. Lu et al. \cite{lu2022assessment} compares the availability of the multi-view information from a single high-overlap image obtained by UAV, the single-view information from the lowest point image, and the mosaic orthophoto, by SVM, ELM, and RF. They find that high-overlapping multi-view images can produce the highest estimation accuracy of leaf and plant N concentrations.

\par \textbf{Biomass estimation:} Biomass is an essential indicator of crop vegetation health and development. However, direct measurement of biomass is expensive and destructive, such as by harvesting, drying and weighing plant samples. RS provides an efficient way to monitor and estimate large areas of AGB due to the diversity of platforms and sensors and the improvement of spatial and spectral resolution. Initially, some traditional statistical models were implemented to link spectral information with the biophysical attributes of crops \cite{chang2005predicting, teal2006season}, but they are always limited by statistical assumptions, failing to address issues, such as outlier data, non-linearity, heteroscedasticity, and multi-collinearity. To improve the modeling of the input-output nonlinear mapping between RS data and biomass, many studies have used various representative algorithms in the field of AI \cite{zhu2019estimating,han2019modeling,bahrami2021deep,gleason2012forest, montes2011high}. For instance, Zhou et al. \cite{zhou2016estimation} proposes a wheat biomass estimation method that combines the vegetation index calculated from China's environmental satellite (HJ) charge-coupled device (CCD) images and RF, and demonstrates the accuracy and robustness of the model at each stage of wheat growth. Bahrami et al. \cite{bahrami2021deep} combines radar and optical earth detection to estimate the leaf area index and biomass of three crops (soybean, corn, and rape) in Manitoba, Canada, using SVM, RF, GBDT, XGBoost models, and a deep neural network. The results suggest that GBDT and XGBoost models have great accuracy in parameter estimation, while the deep neural network has better potential in estimating crop parameters than traditional ML algorithms.

\par \textbf{Weed extraction:} Weeds compete with crops for essential resources such as nutrients, water, sunlight, and space, which can hinder field ventilation and ultimately reduce crop yield and quality. Additionally, weeds can serve as intermediate hosts or habitats for pathogenic microorganisms and pests, leading to the occurrence of diseases and pests in the crop field. Traditional weed detection methods typically calculate multiple vegetation indices based on different bands of information in different growing periods of crops, to analyze the growth status and coverage degree of weeds. However, due to the similarity of spectra between different crops and weeds, this way is often not accurate. As a result, it is crucial to leverage RS technology and AI algorithms to accurately map weeds in fields and develop corresponding measures to remove them. Perez et al. \cite{perez2015semi} proposes a system for generating weed maps using images captured by UAVs. The system utilizes the Hough transform to detect crop row information, improving the accuracy of weed classification. The methods employed include clustering, semi-supervised and supervised learning. Recently, DL methods begin to be employed in the agrifood system. Wang et al. \cite{wang2022weed} introduces an enhanced bionic optimization-based transfer neural network to identify weed density and crop growth. To further improve the identification capability, Reedha et al. \cite{reedha2022transformer} employs the advanced ViT model combined with high-resolution UAV images to classify crops and weeds. These studies demonstrate the advantages of DL models in weed extraction, which can promote the development of precision agriculture. In the appendix, we have shown a schematic diagram of using deep networks for weed detection.

\par \textbf{Crop diseases and pest monitoring:} Crop diseases and insect pests have become the primary factors that reduce agricultural production, causing a large loss in crop yields worldwide. Traditionally, experts have visually identified pathogen and plant disease symptoms, but the accuracy relies on personal experience and is not suitable for large-scale detections \cite{mahlein2016plant}. Nowadays, RS technology for pest monitoring mainly relies on empirical models and ML algorithms \cite{zhang2012detecting, tian2021spectroscopic}. The empirical model is relatively simple, but it is vulnerable to external conditions and lacks universality. ML methods take into account training errors and generalization abilities, overcoming the affection of environment changes \cite{badnakhe2018evaluation}. Lu et al. \cite{lu2017field} collects hyperspectral data of three types of strawberry plants with mobile platforms. They utilize 32 spectral vegetation indices to train stepwise discriminant analysis, fisher discriminant analysis, and KNN algorithms. While Shin et al. \cite{shin2020effect} applies three feature extraction methods (histogram of oriented gradients (HOG),  speeded-up robust features (SURF), and gray level co-occurrence matrix (GLCM)) and two supervised learning classifiers (ANN and SVM) to detect strawberry powdery mildew. They find that ANN and SURF have the highest detection accuracy. These examples demonstrate the potential of ML models in detecting crop diseases and pests.

\par In summary, ML and DL technologies have demonstrated the advantages of high accuracy, strong robustness, and distinguish adaptability for agrifood growth monitoring. By building a monitoring and warning system, effective supervision and diagnosis of crops can be realized, thereby providing real-time, accurate, intelligent, and automatic solutions for remote dynamic management of crop growth, and improving agrifood growth environments as well as increasing agricultural productivity.

\subsubsection{Agrifood yield prediction}

Agrifood yield prediction is a crucial and challenging task in precision agriculture. Typical crop yield prediction refers to the estimation of future productions based on historical data. Given the rapidly increasing global population and the frequency of extreme weather events, efficient and accurate yield prediction is critical for developing food policies and ensuring the security of food supply \cite{alexandratos2012world,lobell2014greater}. Correspondingly, government, dealers, and farmers can also make strategic decisions based on the prediction results \cite{macdonald1980global}. Table~\ref{yielddata} lists the existing available yield prediction datasets and their main characteristics.

\begin{wrapfigure}[]{r}[0em]{0.5\textwidth}
  \centering
  \includegraphics[width=\linewidth]
  {{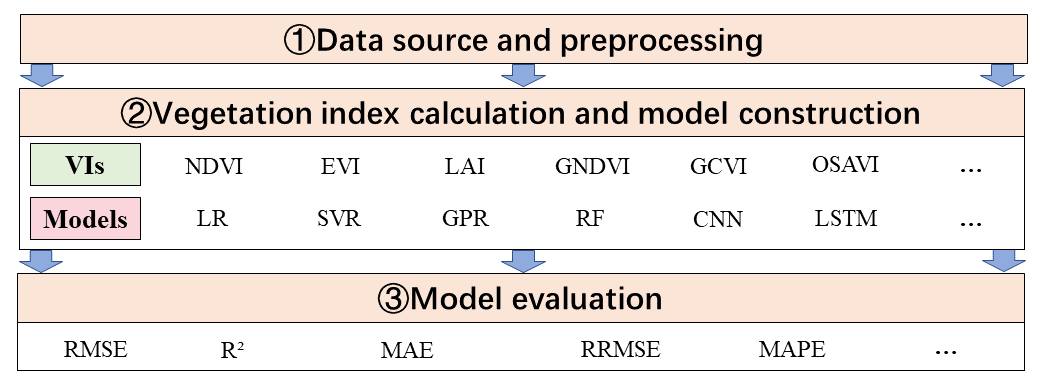}} 
  \caption{The key components of yield prediction.}
  \label{yieldprediction} 
\end{wrapfigure}

\par Current mainstream methods for predicting spatial crop yields can be categorized into two groups: crop simulation models based on growth processes, and statistical ML models. Physical crop models are developed to estimate yield by simulating crop growth and environmental effects such as crop density, light use efficiency, soil nutrient content, and water balance, based on crop growth processes and physiological characteristics \cite{zhang2019california}. Although physical crop models are well-suited to quantify potential yields, they require large amounts of field data related to biotic and abiotic factors for model calibration. As a result, they lack the ability to provide large-scale estimates of actual yields \cite{kastens2005image,sakamoto2013modis}. Statistical ML models do not simulate the biophysical processes of crops but instead, predict yields by establishing a mapping between yield-related factors and historical yield records. This approach allows yield predictions without relying on complex crop growth parameters \cite{wang2020combining}. The key components of statistical ML-based yield prediction have been shown in Fig. \ref{yieldprediction}. Additionally, biophysical indicators that reflect crop growth, such as AGB or net primary productivity (NPP), also can be incorporated as the model input \cite{doraiswamy2004crop,doraiswamy2005application}.

\par By feeding the ML model with the vegetation indices of RS images, such as the NDVI and EVI, we can realize the yield prediction on a large scale.  Linear models are the most basic ML methods. For instance, Johnson et al. \cite{johnson2016crop} inputs various combinations of MODIS-NDVI, MODIS-EVI, NOAA-NDVI, and spectral vegetation indices derived from multispectral and 3D point cloud data into MLR models. Mateo et al. \cite{mateo2019synergistic} combines EVI data and vegetation optical depth derived from soil moisture active and passive L-band to predict county-level yields of corn, soybeans, and wheat in the Midwestern U.S. Corn Belt using the regularized linear regression algorithm. Fu et al. \cite{fu2020wheat} constructs a wheat yield prediction framework by combining various vegetation indices and multiple methods, such as LR and MLR. Moreover, some other linear models including OLS \cite{schwalbert2020satellite,wang2020combining}, least absolute shrinkage selection operator (LASSO)\cite{zhang2019combining,meroni2021yield,zhou2022integrating}, and PLSR \cite{marshall2022field,maimaitijiang2020soybean} have also been used for yield prediction.

\begin{table}[t]
    \centering
    \tiny
    \caption{Summary of available yield prediction datasets.}
    \resizebox{\linewidth}{!}{
    \begin{tabular}{c c c c}
    \bottomrule
        Dataset & Crop & Characteristics & Dataset source \\ \hline

        Barley Remote Sensing Dataset & {\makecell[c]{barley rice, maize, and \\flue-cured tobacco}} & {\makecell[c]{Adopts segmentation label maps with pixel-level \\category annotations.}} & {\makecell[c]{https://tianchi.aliyun.com/dataset/74952}} \\
        
        2018 Syngenta Crop Challenge & corn & 
        
        {\makecell[c]{Contains genetic information, and soil and weather \\variables corresponding to yield performance.}}
        
        & {\makecell[c]{https://www.ideaconnection.com/syngenta-crop\\-challenge/challenge.php}}  \\ 
        
        Global Dataset of Historical Yields (GDHY) & {\makecell[c]{maize, rice, \\wheat and soybean}} & {\makecell[c]{Annual time-series data of 0.5-degree grid-cell on \\yield estimation for major crops worldwide.}}& https://search.diasjp.net/en/dataset/GDHY\_v1\_2 \\ \bottomrule
\label{yielddata}    
\end{tabular}
}
\end{table}

\par Although linear models are able to perform efficient yield prediction, the complex relationships between yields and some factors, such as crop genetics, environmental factors, and human management practices \cite{ray2015climate}, are not always linear. In such cases, linear models may not perform well. To tackle these issues, various nonlinear ML models have been developed to predict crop yields. For instance, Kamir et al. \cite{kamir2020estimating} constructs multiple ML models, including Cubist, MLP, support vector regression (SVR), GPR, KNN, and Multivariate Adaptive Regression Splines, to predict wheat yield in the Australian wheat belt, and finds that SVR achieves the highest prediction accuracy. In another study, Han et al. \cite{han2020prediction} develops a yield prediction model framework using the GEE platform to test the performance of KNN, NN, DT, SVM, and GPR models for winter wheat production in China. Meroni et al. \cite{meroni2021yield} uses SVR-Linear, SVR-RBF, and MLP models simultaneously to predict the national production of barley, soft wheat, and durum wheat in Algeria during the current season. Other commonly used ML models, such as ANN \cite{fu2020wheat}, bayesian neural networks (BNN) \cite{johnson2016crop,ma2021corn}, have also been used to improve the accuracy of crop yield prediction. Nevertheless, since these methods rely on a single model, they may suffer from an overfitting issue when training data is limited.

\par EL techniques have become increasingly popular in recent years to improve the generalization ability of ML models. By combining multiple base models, EL methods can compensate for errors and deficiencies in individual models, leading to more robust and accurate predictions \cite{zhang2019california}. Bagging and Boosting are the two most commonly used EL techniques in ML. Among bagging approaches, the RF employing multiple decision trees  has been widely adopted for crop yield prediction \cite{filippi2019approach,kayad2019monitoring,maimaitijiang2020soybean,han2020prediction}. Filippi et al. \cite{filippi2019approach} develops yield prediction models for wheat, barley, and oilseed rape based on MODIS-EVI time series data and the RF algorithm, respectively. The cross-validation results demonstrate that the RF model yields accurate predictions. In contrast to Bagging methods, many variants of Boosting techniques have also been applied in crop yield prediction. For instance, Zhang et al. \cite{zhang2019california} integrates vegetation indices, canopy cover, and climate data obtained from Landsat and UAV imagery to predict almond yields in the Central Valley of California, using Stochastic Gradient Boosting. Additionally, GBRT \cite{meroni2021yield}, light gradient boosting machine (LightGBM)\cite{zhang2021integrating}, and adaptive boosting (AdaBoost) \cite{wang2020combining} have also been applied in the context of yield prediction. 

\par However, the ML algorithm still suffers from limited representation capability. These issues affect their performances, especially in large-range prediction tasks. Recently, by automatically extracting the inherent deep features of targets, DL technology presents powerful representation ability \cite{lecun2015deep}. It has exhibited superior performances compared to existing process-based crop simulation models or ML methods on large-scale crop yield prediction tasks yield \cite{zhang2021integrating,nevavuori2019crop, maimaitijiang2020soybean}, where CNN and LSTM are the most commonly utilized architectures. Nevavuori et al. \cite{nevavuori2019crop} proposes a yield prediction model for wheat and barley based on 2D-CNN architecture and UAV data, achieving reliable prediction accuracy. Nonetheless, yield prediction often requires the processing of time-series data, as it can be used to capture yield trends as well as cyclical patterns for more accurate predictions. To this end, Fernandez et al. \cite{fernandez2021rice} create 3D-CNN models that simultaneously consider temporal, spectral, and spatial information to predict rice yield in the eastern Arkansas and Terai regions of Nepal, respectively. Here, spectral information reflects the growth condition and physiological characteristics of crops by recording their reflectance in different bands, while spatial information can obtain information on the spatial distribution and growth environment of crops. In addition, some crop growth cannot be fully captured by RS images since real-world crop growth is determined by many factors, such as soil, climate, and human activities. In this case, it may be insufficient to mine deep spatial and spectral features within the image based on CNN alone. To further integrate spatial-spectral depth features and spatial consistency, Qiao et al. \cite{qiao2021exploiting} combines multi-kernel learning techniques to integrate two types of heterogeneous information into a 3D-CNN model to improve the yield prediction accuracy in three types of winter wheat growing areas. In addition to CNNs, the LSTM model, as a variant of the recurrent neural network, is particularly well-suited for processing temporal data due to the use of hidden states to capture information from previous states. In \cite{zhang2021integrating}, Landsat imagery and MODIS imagery are combined into the LSTM for yield prediction in maize-growing areas of China, where the LSTM outperforms LASSO and LightGBM algorithms. Similarly, Schwalbert et al. \cite{schwalbert2020satellite} uses multivariate OLS linear regression, RF, and LSTM models to predict soybean yields in southern Brazil, using a combination of MODIS multi-class products.

\begin{figure*}[t]
        \centering
        \includegraphics[width=0.8\linewidth]{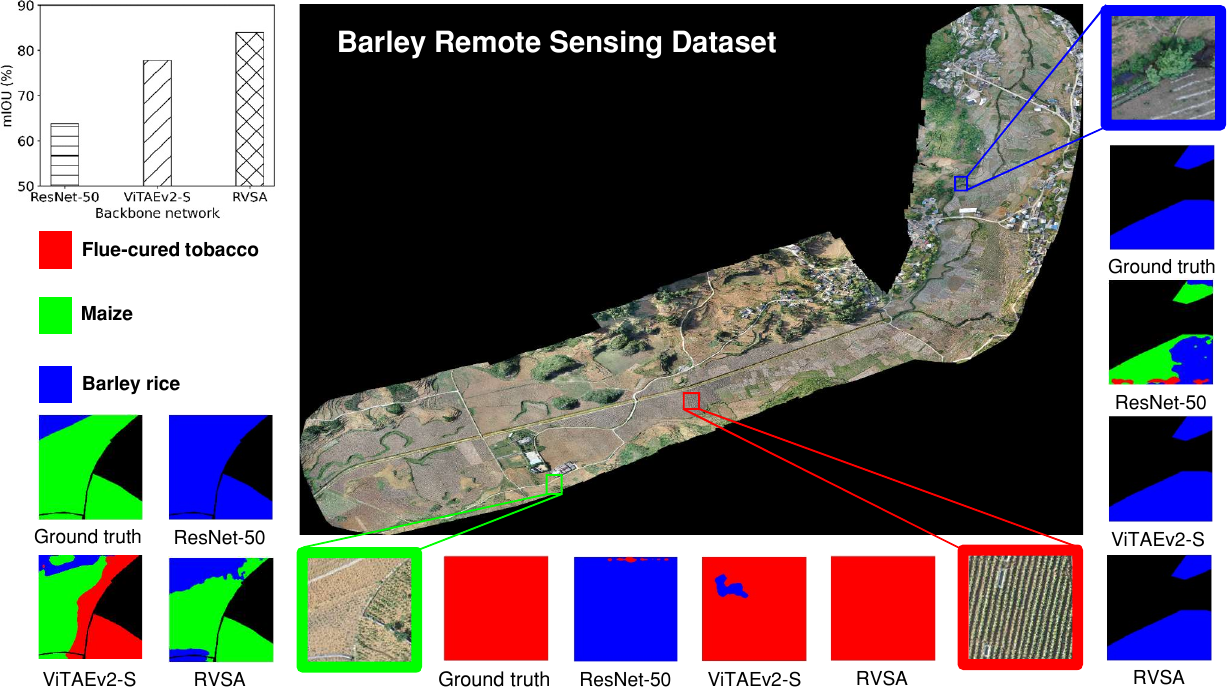}   
    \caption{Segmentation results on Barley Remote Sensing Dataset using ResNet-50 \cite{he2016deep}, ViTAEv2-S \cite{zhang2023vitaev2} and RVSA \cite{wang2022advancing} as the backbone of UperNet \cite{upernet}, respectively.}
    \label{rs_seg_agri}
\end{figure*}

\par Several recent studies have explored the feasibility of using spatiotemporal architectures that simultaneously contain CNN and LSTM to process RS time series data for crop yield time series modeling and prediction. In \cite{jeong2022predicting}, the performances of five different network structures ((FFNN), 1D-CNN, LSTM, 1D-CNN+LSTM, and LSTM+1D-CNN hybrid models) for the early prediction of rice yield are evaluated. In these models, the CNN module is used to extract spatial features, and the LSTM module is used to obtain sequence features. Both hybrid models outperform the single models. Nevavuori et al. \cite{nevavuori2020crop}  has indicated that CNN and LSTM can be connected and combined in an improved manner, where CNN-LSTM, ConvLSTM, and 3D-CNN architectures are developed for wheat, barley, and oat yield prediction. Wang et al. \cite{wang2020winter} creates a two-branch network to predict winter wheat yields at the county level in major production areas of China, where the first branch composes of an LSTM network with meteorological and RS data inputs, and the other branch exploits CNNs to model static soil features. Qiao et al. \cite{qiao2021crop} constructs a spatial-spectral-temporal neural network by connecting the top of a 3D CNN and several stacked bidirectional LSTM units. They predict winter wheat and maize yields in three regions of China using multispectral and multitemporal MODIS images. We have summarized related works on crop yield prediction using RS images with AI technologies, as shown in the appendix.

\par  Besides classification, in our consideration, segmentation models also have the potential to be used for yield prediction. However, according to our literature review, existing related research has not yet utilized segmentation networks. For this purpose, on the barley RS dataset (see Table~\ref{yielddata}), we utilize the UperNet \cite{upernet} as the segmentation framework to segment the regions of flue-cured tobacco, maize, and barley rice, where three different backbone networks including ResNet-50 \cite{he2016deep}, ViTAEv2-S \cite{zhang2023vitaev2} and RVSA \cite{wang2022advancing} are employed. To reduce the domain gap and improve performance, ResNet-50 and ViTAEv2-S are pre-trained on a large-scale RS dataset following \cite{wang_rsp_2022}. We show the segmentation results in terms of the mean intersection over union (mIOU) in the upper-left of Fig. \ref{rs_seg_agri}. Compared to classical ResNet-50, the models with advanced vision transformer structures deliver higher accuracy. Among them, RVSA performs the best since it is designed specifically for RS tasks by considering the characteristics of RS targets. Fig. \ref{rs_seg_agri} also provides the segmentation maps of three typical areas, demonstrating that the advanced vision transformer model generates fewer misclassified results. The segmentation map provides a direct measure for area estimation, which is useful for large-scale yield prediction.

\par Currently, since spatiotemporal reference data based on pixel-level yields is challenging to obtain, most studies mainly focus on the prediction at large scopes, where the county is the smallest scale. Fewer works are conducted at smaller scales, e.g., the farm scale. Nevertheless, in order to fully understand the influencing factors and mechanisms related to crop yields, conducting experiments on farm-scale and pixel-level yield may be more useful than county-scale, and related topics need further exploration in the future.

\par In addition to RS images, AI technologies also have the potential for agrifood yield prediction through natural images. To this end, we evaluate the performance of Sparse R-CNN detection framework \cite{sparse_r_cnn} on the global wheat detection dataset\footnote{https://www.kaggle.com/competitions/global-wheat-detection/overview}, more experimental details can be found in the appendix.

\begin{wrapfigure}[]{r}[0em]{0.45\textwidth}
    \begin{minipage}{1\linewidth}
        \centering
        \includegraphics[width=\linewidth]{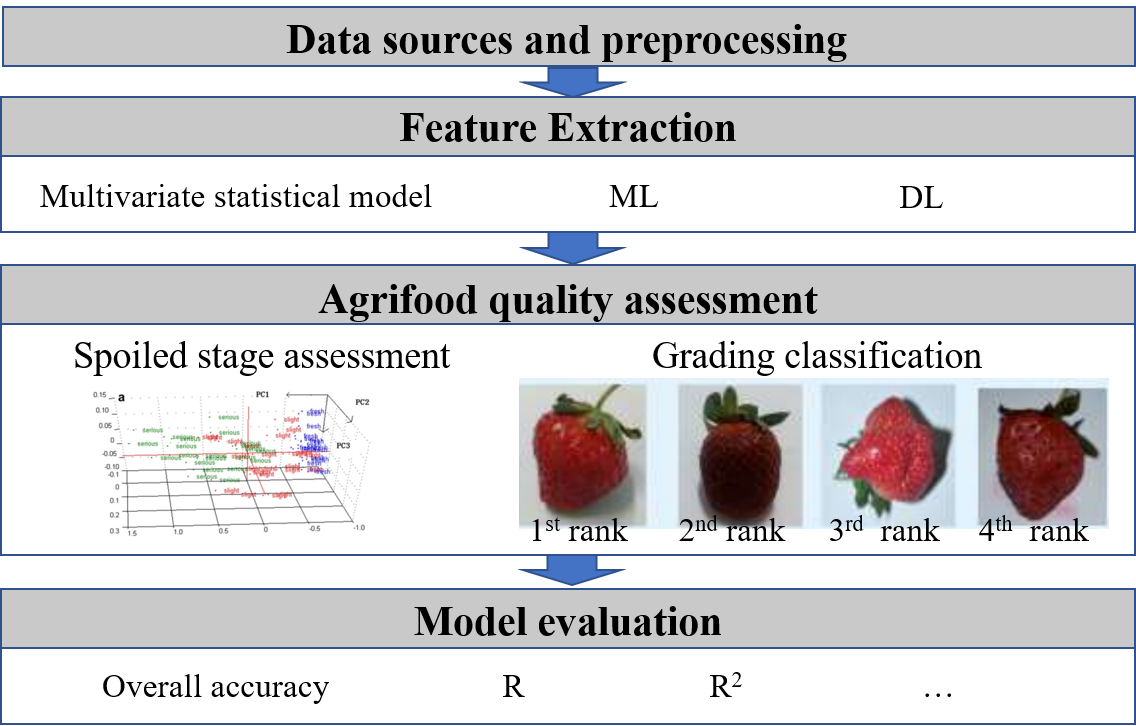}
    \end{minipage}    
    \caption{The key components of agrifood quality inspection.}
    \label{agrifood_quality}
\end{wrapfigure}

\subsubsection{Agrifood quality assessment}

\par Crop quality assessment involves evaluating various quality attributes of crops and grading them accordingly. ML techniques are highly relevant for both tasks as they have an excellent capacity to handle multidimensional data by incorporating several predictor features. Fig. \ref{agrifood_quality} depicts the key components of the agrifood quality assessment process by taking the example of strawberries. In the existing literature, Dong et al. \cite{dong2013analyzing} uses long-range Fourier transform infrared spectroscopy to capture the spectral characteristics of volatile organic compounds (esters, alcohols, ethylene, etc.) generated by strawberries after different storage times for detecting abundance changes. Then, they perform PCA to distinguish fresh, slightly spoilage, and spoilage categories. Mahendra et al. \cite{mahendra2018comparison} compares seven types of features and uses SVM to classify fruits into two groups: good and damaged. Recent developments have shown that DL models, compared to traditional ML models, perform better in fruit grading due to their ability to extract high-level features from raw input data. Sustika et al. \cite{sustika2018evaluation} evaluates the ability of CNN-based architectures (e.g., AlexNet \cite{krizhevsky2017imagenet}, GoogLeNet \cite{szegedy2015going} and VGGNet \cite{chollet2017xception}) for binary classification and grading strawberry fruit into four levels using RGB images. In a word, AI techniques have demonstrated their potential in the quality inspection of agricultural products, leading the industry towards intelligence and automation.

\subsection{Artificial Intelligence Methods in Animal Husbandry}

\subsubsection{Pasture monitoring and evaluation}

Grassland plays an important role in maintaining the ecological system \cite{fernandez2022estimating}. Due to the contribution of carbon storage, grasslands are mainly used for animal feed production. The yield of pasture directly affects animal husbandry production. Therefore, the countries with a developed animal husbandry industry typically have vast grassland areas. Relevant monitoring is important for the utilization and management of pastures, and the improvement of ecological environments \cite{xu2008modis}.

\par Accurately and effectively modeling the above-ground biomass (AGB) of grasslands is a crucial and fundamental task for monitoring and managing grasslands in pastoral areas. AGB can be estimated using both traditional (or ground-based) and RS methods. Traditional methods include visual inspection, cutting and drying, plate lifter, and in situ spectroscopy. However, these methods are time-consuming and only applicable in small-scale monitoring \cite{xu2008modis}. 

\par With the advancement of RS temporal and spatial resolutions, large-scale, efficient, and real-time monitoring of pasture biomass has become in reality.  In the past four decades, many methods have been developed for estimating pasture biomass based on satellite RS data. These methods can be divided into three main categories: vegetation index-based \cite{hill2004estimation}, biophysical simulation \cite{johnson2008dairymod} models, and ML algorithms. For instance, Rosa et al. \cite{de2021predicting} tested the performance of an integrated method combining multispectral RS imagery obtained from UAV, statistical models (generalized additive model, GAM), and ML algorithms (RF) to predict future pasture biomass loads. The study shows that using observations of pasture growth along with environmental and pasture management variables enabled both models, i.e., GAM and RF, to predict the pre-grazing pasture biomass at field scale, with an average error of less than 20\%. Chen et al. \cite{chen2021pasture} developed a sequential neural network model by utilizing Sentinel-2 time-series data, weekly field biomass observations, and daily climate variables to estimate the pasture biomass in dairy farms, obtaining a relatively high prediction accuracy. Rangwala et al. \cite{rangwala2021deeppastl} developed a DL-based pasture prediction model (DeepPaSTL) to estimate pasture biomass by predicting the sward heights of the entire pasture. Their model could perform within a 12\% error margin even during the period of the maximum growth dynamics on the ranch.

\par Biochemical parameters of grassland forages are crucial indicators for measuring vegetation growth status, forage feed value, grass and animal husbandry nutrient balance, and carbon cycling \cite{tong2013progress,gao2019modeling}. One of the key parameters that have been identified as significant for crop growth and yield is P, due to its essential role as a source of nutrients. Inadequate availability of this nutrient has been observed to have a detrimental effect on crop growth and yield. To this end, Gao et al. \cite{gao2019modeling} constructs 39 models using hyperspectral data and multiple factors to estimate the P content of the grasslands in the eastern alpine region of the Tibetan Plateau.

\subsubsection{Animal individual monitoring} 
In addition to monitoring animal husbandry behavior at the group level using remote sensing images, modern animal husbandry also requires monitoring individual animals to ensure their health and growth and prevent the spread of disease.

\textbf{Animal individual recognition:} To conduct individual animal monitoring in animal husbandry, it is necessary to recognize and distinguish each animal on the farm. This recognition is achieved by identifying subtle differences between individuals, such as the length of a cow's face, the shape of a sheep's paw, or the patterns on a dairy cow's coat. Early methods for individual recognition~\cite{hansen2018towards} involve directly capturing images of the animals and using CNNs for recognition. However, these methods may not be suitable for deployment on edge devices due to the heavy computational load. To address this issue, a lightweight CNN has been introduced in \cite{li2021individual} for easy deployment and fast inference. To further improve recognition accuracy, some methods require the model not only to recognize individual animals but also to distinguish their postures~\cite{yu2ap} or segment them from the background~\cite{yang2018high,zhu2020automatic}. Moreover, for specific animal species, distinct parts of the animals, such as the face of pigs, can be utilized for better recognition~\cite{hansen2018towards}.

\textbf{Animal sickness detection:} Monitoring the health of individual animals is crucial for ensuring their growth in animal husbandry. Facial expression is a general indicator of an animal's discomfort, and automated recognition of facial expressions can help detect sickness. For instance, Noor et al. \cite{noor2020automated} uses facial expression analysis of sheep to evaluate their health status by feeding their facial images into CNNs. However, capturing the facial images of each individual can be challenging and time-consuming. To overcome this issue, several methods~\cite{wang2020cattle} use images captured from the front of the animals as input and require the networks to jointly detect the animal's face location and evaluate their health status. In addition to evaluating an animal's health status based on facial expression, specific techniques are designed to diagnose particular illnesses. For instance, to diagnose lameness, some methods~\cite{wu2020lameness} first detect and track animals through a series of contiguous frames, then collect their trajectory for analysis. To improve the accuracy of diagnosis, it is beneficial to estimate the poses~\cite{yu2ap,xu2022vitpose} and their temporal correspondence based on optical flow~\cite{xu2022gmflow}. Furthermore, Van et al. \cite{van2018implementation} adopts 3D convolutions to process contiguous frames for better analysis of temporal information. AI technology has also been used to predict the spatial incidence of schistosomiasis in sheep \cite{suresh2022exploration}, where the RF and AdaBoost models are identified as the best models. Apart from using RGB images of individual animals for evaluation, other clues can be utilized to aid diagnosis. For example, thermal images can be used to help identify cow mastitis \cite{xudong2020automatic}, while fecal images can be used to evaluate the health of cattle's digestive systems \cite{owens2016mathematical}.

\textbf{Animal behavior monitoring:} Animals, like humans, exhibit unique behavior patterns that are specific to each individual. Collecting and analyzing these patterns can aid in their growth and well-being. The stacked RGB frames or optical flow maps can be used to recognize different behaviors, such as drinking, feeding, and lying. We have presented some pose estimation examples in the appendix for illustration. AI techniques can also be used to analyze an animal's engagement with various objects for identifying their preferences and improving their mood, and monitoring patterns of aggressive behavior during interactions between individuals to prevent harm \cite{fuentes2020deep,tsai2020assessment}.

\subsubsection{Other topics} 

Automatic techniques of animal product generation are increasingly important aspects of modern animal husbandry. For example, image processing techniques can be used to determine the optimal storage conditions for meat. While Pu et al. \cite{pu2015classification} uses visible and NIR hyperspectral images to determine whether meat is fresh or not. This approach can accurately identify the freshness of the meat, ensuring food safety. In addition, AI methods have been also utilized for monitoring animal husbandry fences and poultry houses \cite{zingman2016detection}.

\par From the above literature review, it can be seen that various AI technologies including statistical ML and DL methods have been improving the animal husbandry system from diverse perspectives. Notably, we conduct a statistical analysis of various AI models used in animal husbandry research across 69 articles, among them, about 30\% of the articles use the RF model, 25\%  of them adopt SVM and ANN models, and about 10\% of them resort to deep learning models.

\subsection{Artificial Intelligence Methods in Fishery}

\subsubsection{Fishing area identification and prediction}

\par The identification and prediction of fishing areas have become increasingly important in the era of large-scale fishery exploitations. It is beneficial for ensuring the safety of fisheries, forecasting fishing opportunities, assessing fisheries resources, and managing fishing operations. In the appendix, we have summarized typical works on fishery-related tasks with AI algorithms.

\par The scale of the farming area is a critical factor in determining fishery production \cite{cheng2020research,zeng2019extracting,ottinger2017large}. Obtaining a quick and accurate estimation of the large-scale regional farming area can provide an overall understanding of the farming status. To achieve this, Cheng et al. \cite{cheng2020research} proposes a new semantic segmentation network, i.e., the hybrid dilated convolution U-Net (HDCUNet), which combines U-Net with hybrid expanded convolution to extract coastal aquaculture areas from GF-2 images. This approach solves the issue of misclassifying floating objects on the water surface as aquaculture areas and achieves an overall accuracy of 99.2\%. In recent studies, with the RS images, researchers have focused on the geometric features to extract aquaculture pond areas \cite{zeng2019extracting, zeng2020rcsanet}. Concretely, Zeng et al. \cite{zeng2019extracting} involves extracting water surfaces from satellite images and performing boundary tracking for each water segment. The geometric features such as perimeter, curvature, and contour-based regularity of the water surface objects are then evaluated using SVM to extract aquaculture ponds. Zeng et al. \cite{zeng2020rcsanet} also considers the same directional correlation between pixels of aquaculture ponds. For this purpose, based on the extracted features by the FCN, they separately use row and column self-attentions to extract aquaculture ponds from high spatial resolution RS images. In addition to optical RS images, other waveband data such as SAR images are also leveraged in some studies. For example, Ottinger et al. \cite{ottinger2017large} extracts aquaculture ponds based on backscatter intensity, size, and shape features.

\par The rational increase of potential fishing areas \cite{ liang2021semi,sivasankari2022he,shi2018automatic} is a crucial factor in maintaining the continuous growth of marine fisheries. To achieve this, Shi et al. \cite{shi2018automatic} proposes an automatic raft-by-pixel labeling method using a fully convolutional dual-scale network structure that captures intricate details without downsampling. Liang et al. \cite{liang2021semi} utilizes pseudo-labels generated by conditional generative adversarial nets to improve extraction accuracy. While Sivasankari et al. \cite{sivasankari2022he} proposes a network structure named hybrid ensemble deepfishnets that integrates deep convolutional networks and filter bat recurrent neural networks, achieving remarkable performance.

\subsubsection{Fish production forecast}

\par Benefiting from the advancement of geospatial observation technology, it is feasible to acquire copious amounts of synchronous and dynamic information regarding fish culture areas, which has enhanced the ability to conduct comprehensive studies using AI technology within the fisheries domain. Environmental data from potential fishery culture areas can also be collected and analyzed qualitatively to determine the distribution characteristics of fish organisms. These data can be used to support yield prediction and optimization of fish products. Knudby et al. \cite{knudby2010predictive} combines the habitat variables from IKONOS data with ML and statistical models to estimate the species richness and biomass of fish communities around two coral reefs in Zanzibar. Li et al. \cite{li2021seecucumbers} uses the object detector YOLOv3~\cite{2018YOLOv3} to detect sea cucumbers from UAV images of Hideaway Bay, Queensland, Australia, providing the first example of applying a DL model to quantify the number and density of sea cucumbers over a large area. Besides RS data, other types of data have also been utilized in fishery research. Zhang et al. \cite{zhang2022coastal} creates a benchmark of a large-scale underwater video dataset for training the mask R-CNN model. Chen et al. \cite{chen2022remote} adopts ML algorithms to simulate the effects of environmental variables on fish catches, finding that the catches of aquatic species in Poyang Lake are highly susceptible to hydrometeorological conditions. Besides, real-time detection techniques have been developed to facilitate the capture and monitoring of underwater fish or crabs \cite{cao2020real, ji2023real}. Yao et al. \cite{yao2024fmrft} proposed an end-to-end real-time tracking model to achieve accurate monitoring in factory farming environments. However, the collection and labeling of large-scale datasets is a labor-intensive task. To address this issue, researchers in \cite{mahmood2020automatic} use synthetic data to improve the accuracy of detecting western rock lobsters.

\subsubsection{Fish product classification} Unlike crops and animal husbandry products, fishery products are typically alive and cannot be easily labeled using tools like radio frequency identification. As a result, supply chains of fishery products rely on individual recognition techniques, such as those used for lobsters \cite{vo2020application}. In addition, CNNs are employed to distinguish between different fish categories \cite{monteiro2023fish}.

Overall, these studies showcase the potential of utilizing various types of data and AI techniques in fishery research for a better understanding and management of aquatic resources.

\section{Challenges}

Despite the great potential of AI technology in agrifood systems, challenges still exist. In this context, we summarize five challenges related to agricultural characteristics, external factors, data acquisition and processing, model design and maintenance, and ethical risks.

\noindent\textbf{Agricultural characteristics} Agricultural production has unique characteristics such as regionality and seasonality. These characteristics pose new challenges for the application of AI in agriculture. For example, the long growth cycle of crops can limit the real-time performance of technologies, while the variability in farming practices introduces further complexity. Advanced AI models need to consider the wide range of agrifood products produced in different regions and incorporate these diverse agricultural characteristics to enhance their predictive capabilities and robustness. Utilizing region-specific data and developing localized models may be helpful to address these challenges.

\noindent\textbf{External factors} Agricultural activities are affected by multiple natural conditions, such as water resources, soil nutrients, terrains, and climates. Climate factors, including light and precipitation, are considered major impact factors on agrifood industries. The complexity and variability of these factors make it difficult to develop adaptable AI models. The intensification of climate change, especially in regions with extreme or unpredictable weather, further poses challenges for applying AI technologies. Developing transfer learning and domain adaptation techniques can possibly improve model robustness across various environments. Additionally, integrating real-time environmental data and IoT devices is available to enhance the accuracy of AI predictions in changing conditions. Furthermore, some adverse conditions, such as weak network infrastructure in rural areas of developing countries, can limit the integration of AI, IoT, and agriculture. Therefore, it is essential to improve the construction of network infrastructure in rural areas, promoting the deployment of AI technologies in agrifood systems.

\noindent\textbf{Data acquisition and processing} Agricultural practices are long and complex processes, and relying solely on a single data resource may not lead to effective judgments and expected outcomes. In the agrifood system, data is often heterogeneous as it is obtained from various sources and stored in multiple databases with diverse formats, following a series of preprocessing procedures (see Sec.~2.2.2 for details). The scarcity of high-quality annotated data further complicates this issue. To address this challenge, employing multiple data-centric techniques, such as data augmentations, data synthesis, and developing collaborative data-sharing platforms may improve data acquisition and integration processes. In addition, the utilization of advanced multimodal data fusion methods also has the potential to enhance the quality and efficacy of the collected data. For example, data fusion approaches and machine learning algorithms can be adopted to effectively process and analyze heterogeneous data collected by UAVs and handheld sensors.

\noindent\textbf{Model design and maintenance} Designing algorithms with a large amount of data for complex agrifood systems is typically a time-consuming process. Correspondingly, the resulting AI models can be sophisticated and have high computational complexity, making deployment difficult for farmers. For these issues, more efficient model design techniques, such as automated machine learning, can be used to reduce the time and complexity involved in developing AI models. In addition, applying lightweight models and leveraging edge computing can mitigate deployment challenges, while continuous and incremental learning strategies enable models to be up-to-date with minimal computational overheads. Moreover, regular maintenance and updates of AI models are crucial to adapt to new data and changing conditions. Therefore, it is necessary to develop user-friendly interfaces and conduct training for farmers, benefitting effective usage of AI technologies.

\noindent\textbf{Ethical risks} AI solutions are frequently employed in critical decision-making scenarios that involve access to private data. Improper use of such data can result in various legal issues, such as privacy breaches and malicious attacks. Lack of transparency can undermine the trust of farmers and consumers, leading to their reluctance to adopt AI solutions and hindering the development of AI technologies. To tackle these issues, AI systems can be designed with privacy-preserving techniques, such as differential privacy and secure multi-party computation. The explainable AI techniques may be useful in enhancing transparency during AI decision-making processes and building trust among users \cite{explain_agri_ai_1,explain_agri_ai_2,explain_agri_ai_3,explain_agri_ai_4}. In addition, establishing clear data governance policies can also mitigate ethical risks, which are essential to protect sensitive data and maintain user trust. The widespread implementation of AI technology in agrifood systems may raise concerns about unemployment issues, especially for easily automated jobs, such as planting and pest control. Nevertheless, the AI-driven agrifood systems may also create new job opportunities in data analysis, system monitoring, and model maintenance, highlighting the need for reskilling to adapt farmers to new occupational roles for better leveraging the benefits of AI methods. The stakeholders can be also engaged in the model development process to further ensure that AI solutions align with ethical standards and meet the needs of the agricultural community.

\section{Opportunities}

The deployment of AI technology in agrifood systems has also activated many opportunities, where cutting-edge exploration experiences in AI fields, such as multimodal and scalable models can be leveraged by agriculture applications, improving software and hardware infrastructures simultaneously. In this section, we attempt to introduce several promising development directions for agrifood systems after adopting AI techniques, including foundation models, Trustworthy AI, and IoT technologies.

\subsection{Foundation Models Contribute Modern Agricultural Systems }

\subsubsection{Large-scale data pre-training} Recently, large foundation models have attracted increasing attention due to their great potential (e.g., high data efficiency) across various domains \cite{liu2022swin,zhang2023vitaev2,chen2022pali, wang2022advancing}. Typically, these models are pre-trained on large-scale data~\cite{dosovitskiyimage}, either supervised or unsupervised, and possess exceptional generalization ability due to their extracted representative features. As a result, they can adapt well to various tasks by fine-tuning with only a fraction of task-specific data. For instance, the pre-trained chatGPT model \cite{ouyang2022training,brown2020language} can perform translation or math tasks with just a few prompt sentences. To more intuitively demonstrate the capabilities of large models, we finetune different ImageNet pre-trained models on the grocery store dataset \cite{grocery_store_dataset}, which is used for agrifood image recognition, more details have been shown in the appendix. The experimental results indicate that, the desirable property of large models reduces the need for collecting and annotating a large amount of data in agrifood systems while achieving better performance. Additionally, the long-term cycle and large-span properties of agrifood data provide a wealth of information that can be leveraged for pre-training large models, making them well-suited for use in agrifood systems.

\subsubsection{Multi-modal approaches} Foundation models possess an important property that allows them to use multiple types of data jointly to achieve more accurate predictions~\cite{radford2021learning}. Agrifood data comes from various sources, such as satellites, UAVs, and other sensors. However, current approaches typically rely on a single data type as input and overlook the complementary between diverse data sources. Foundation models have the potential to learn from these data jointly, which could improve prediction accuracy for tasks such as crop yield prediction or pasture monitoring. By using visual images from UAVs, RS images from satellites, and soil and climate conditions from sensors as inputs, foundation models can provide more accurate and comprehensive results.

\subsubsection{Multi-task networks} Current methods tend to design one method for each specific task. Although they have obtained superior performance in agrifood systems, such a modeling pipeline introduces extra deployment costs as we need to design and adapt different models for different tasks. It also should be noted that foundation models have the potential to unify different tasks using a single model~\cite{yuan2021florence}. For example, (OFA)~\cite{wang2022ofa} jointly model detection, segmentation, and (VQA) tasks within one model. Such a joint modeling pipeline can help the foundation model efficiently leverage the labeling information provided by different tasks and obtain better performance in different tasks.

\subsubsection{Cross-domain foundation models} Except for the applications discussed in this paper, the large-scale foundation models can provide more opportunities for a better agrifood system. For example, large-scale models can be used to predict future climates and help farmers make better decisions regarding the harvesting and seeding time~\cite{ravuri2021skilful}. The medical foundation model contributes to the discovery of better fertilizer and helps identify sick animals. The biology-related foundation models~\cite{cramer2021alphafold2} may help the researchers by alleviating their cost of finding better hybrid plants by predicting the properties of specific hybrid plants given the parent plants growth situation.

\subsection{Trustworthy AI Technologies Rebuild Our Agrifood systems}

\subsubsection{Safe and traceable technology}
As one of the foundation systems in society, the security and trustworthiness of AI techniques are essential for modern and safe agrifood systems. For example, the techniques like blockchain~\cite{zheng2018blockchain} and smart sensors~\cite{zhang2020empowering} can be used to monitor the growth of agrifood, detect contaminants, and predict food safety risks. They can also be used to enhance supply chain management by providing a secure and transparent record of food production and distribution and reducing the risk of fraud. Text detection and recognition AI models \cite{du2022i3cl, he2022visual,ye2023deepsolo} enable reading product information of agrifood (see in the appendix) and provide means for tracing agrifood. These changes will improve public trust in modern agrifood systems.

\subsubsection{Explainable AI}

Although AI technologies such as DL have greatly benefited the agrifood system and contributed to modern agrifood systems like precision agriculture technology, understanding the decision mechanism behind AI is still worth further exploration. For example, although DL can make accurate predictions about the crop yield, it can not explain which key elements it chooses and how it balances them in making the predictions. Such a black-box property may restrict their further applications in agrifood systems and make it hard for researchers to improve these models. The interpretable AI technologies, instead, focus on revealing the dark secrets behind the AI technologies and help AI make explainable decisions regarding the key elements and the logic it considers in prediction \cite{explain_agri_ai_1,explain_agri_ai_2,explain_agri_ai_3}. With such interpretable AI technologies, users can gradually put more trust in the AI systems with informed decisions and improve the prediction results, establishing a trustworthy relationship between the model and the user.

\subsubsection{Robust models} Apart from the above opportunities, the robustness of the AI models is also an important topic in agrifood systems. Due to the influence of weather change or location variants, the captured data may have significantly different distributions from the data used in training. Making the models robust to data with different distributions can greatly simplify the deployment of AI technologies in agrifood systems and improve the reliability of the prediction results made by AI \cite{rubust_agri_ai_1, rubust_agri_ai_2}.

\subsubsection{Resilience and Sustainability} The development of AI methods can improve energy management and promote the resilience and sustainability of agrifood systems \cite{cubric2020drivers, rejeb2022examining}. AI enables precise modeling of energy requirements for farming activities, thereby promoting energy conservation and the production of environmentally friendly agricultural goods \cite{bolandnazar2020energy}. For instance, AI methods can be used to strictly control the amount of watering and fertilizing during crop growth to alleviate the current serious problem of water and fertilizer abuse. Moreover, the adoption of AI and the development of smart agriculture can motivate more farmers to utilize green energy sources  and mitigate the carbon footprint associated with farming practices and food production \cite{ragazou2022agriculture}. The agrifood industry also stands to benefit by leveraging AI to optimize energy consumption and enhance productivity with fewer energy resources \cite{mor2021artificial}. Therefore, future research should also focus on developing efficient AI models to predict the optimal amount of energy necessary for agricultural activities \cite{nabavi2018integration}.

\begin{wrapfigure}[]{r}[0em]{0.5\textwidth}
    \begin{minipage}{1\linewidth}
        \centering
        \includegraphics[width=\linewidth]{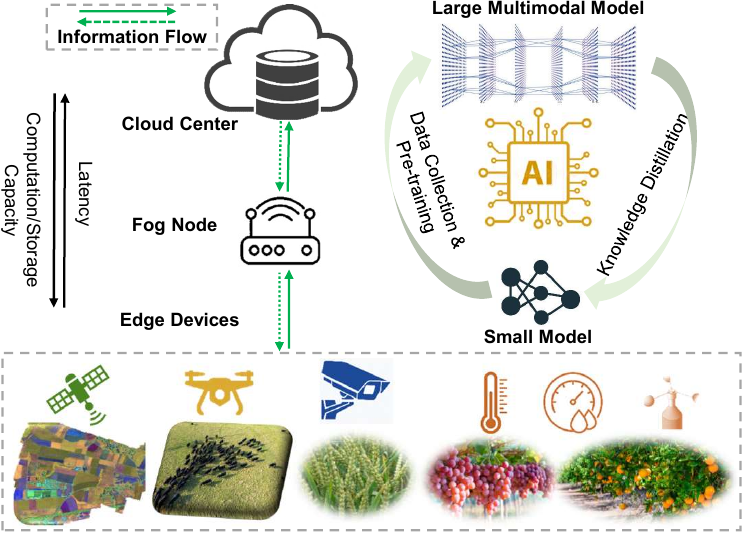}
    \end{minipage}    
    \caption{Combining AI and IoT creates a closed-loop evolution framework for the agrifood system.}
    \label{aiot_foundation_model}
\end{wrapfigure}

\subsection{AIoT Technologies Reshape the Agrifood systems}

Apart from algorithm development, hardware systems have also experienced rapid development. AIoT in the agrifood system can also present several key opportunities \cite{zhang2020empowering}. AIoT possesses a tri-tier architecture, it can be divided into three types of computing layers: cloud, fog, and edge, where the proximity to the data source is gradually reduced. Usually, the computation and data storage capacities of cloud, fog, and edge are sequentially decreased, with continuously reduced latency. Leveraging these characteristics, we can construct a hybrid intelligence computation structure by reasonably assigning computation workloads of different models. Fig.~\ref{aiot_foundation_model} portrays a promising vision of integrating modern AI technologies, including large multimodal foundation models, into the IoT ecosystem. Initially, edge devices collect agrifood data to realize fundamental analyses based on the deployed small model. Then, high-level analyses can be obtained by the large foundation models on the cloud node, which are pre-trained on the collected vast amount of data. The knowledge of large foundation models can be distilled and transferred to smaller models, thereby further improving their capabilities. This process creates a closed-loop evolution framework for the agrifood system. In this way, AIoT systems can provide timely feedback and analysis for the massive data captured from extensive sensors, assisting users to make more informed predictions. For example, advanced AIoT technology can monitor the health of animals or agriculture in real-time, which can be used to further improve food security and contribute to the development of traceable AI in agrifood systems.

\section{Conclusion}
In this survey, we systematically investigate the implications of AI technologies on reshaping the agrifood industry. First, we provide an overview of the agriculture data involved in applying AI models, covering the perspectives of source, storage, and processing. Next, we summarize the existing AI methods, including traditional ML methods and recent popular DL models, which are widely used in agrifood systems. We then conduct a detailed review of various AI applications in diverse agricultural fields, such as planting, husbandry, and fishery. Our focus lies primarily on the utilization of AI for identification, monitoring, and prediction tasks while also covering a range of subtopics within these stages. Finally, we discuss the opportunities and challenges of integrating AI and agriculture fields. 

Through investigating and analyzing existing research, we summarize the following key takeaways: 1) The development of data acquisition technology has brought massive, multi-source, multi-temporal, and refined data, which has been driving agrifood fields to make continuous breakthroughs in extensive applications. 2) Leveraging AI techniques from various fields, such as computer vision and ML, numerous agrifood-oriented methods have been developed in multiple tasks of agriculture, animal husbandry, and fishery to enable the prosperity of the agrifood industry. 3) Although agrifood systems have experienced rapid development through adopting AI technologies, the inherent agricultural production characteristics and external environmental factors, the constraints lying in data and model aspects, and the raised ethical issues, pose challenges for the further progress of agrifood fields. To this end, we point out three promising directions in the future: foundation model, trustworthy AI, and AIoT. The large multimodal foundation model is expected to enhance cross-task transferring performance, providing unified solutions for all scenes. Trustworthy AI is beneficial to enhance security, transparency, and reliability, ultimately building public trust for agrifood systems. AIoT boosts the real-time data analysis capabilities of agrifood systems, allowing users to make more precise decisions.

We hope this survey will uncover the immense potential of AI in agrifood systems, inspire further research and practical applications in related fields, and foster discussions on the ethical use of AI technologies, all with the goal of enhancing the productivity, efficiency, safety, and sustainability of our agrifood systems.

\bibliographystyle{ACM-Reference-Format}
\bibliography{Agrifood}


\clearpage

\section*{Appendix}

\appendix

\section{Artificial Intelligence Methods in Agriculture}

\subsection{Agrifood classification}

\begin{figure}[h]
    \centering
    \includegraphics[width=0.7\linewidth]{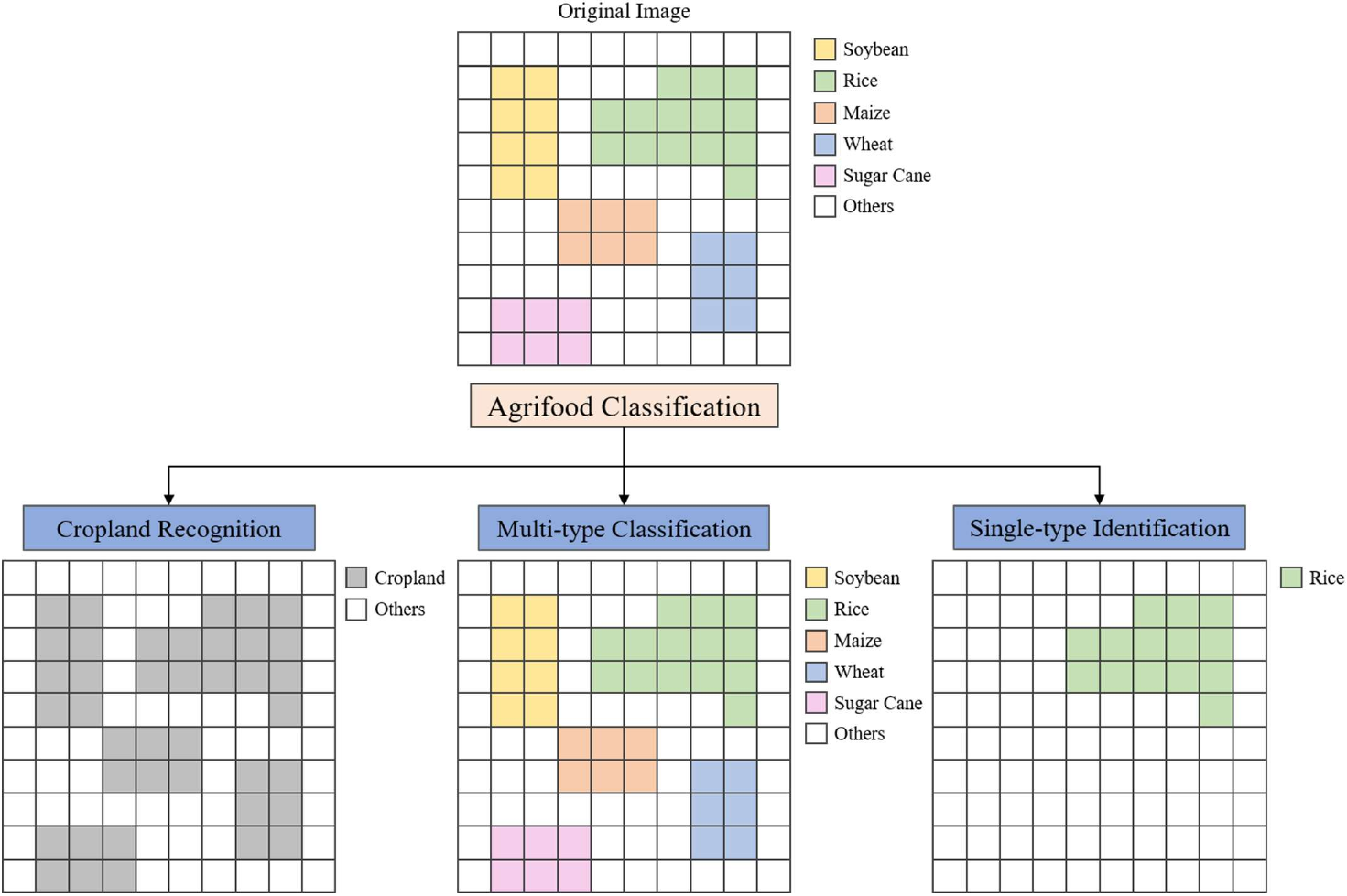} 
    \caption{Illustration of three common agrifood classification tasks.}
    \label{yy_classification}
\end{figure}

As mentioned in the main paper, we have summarized remote sensing (RS)-based agrifood classification tasks into three categories. 
A diagram of these three categories is shown in Fig. \ref{yy_classification}.

\subsection{Agrifood growth monitoring}


Table \ref{tablegrowth} lists typical literature related to agrifood growth monitoring involving RS and machine learning (ML) algorithms.

\begin{table*}[h]
\centering
\caption{Summary of agrifood growth monitoring methods based on remote sensing and machine learning.}
\resizebox{\linewidth}{!}{
\begin{tabular}{p{3cm}<{\centering}p{1cm}<{\centering}p{2cm}<{\centering}p{11.8cm}p{1cm}<{\centering}}
\hline
Author    & Year & Field & \makecell[c]{Description}                                                                                                                                                                                                                                                          & Reference \\ \hline
Babaeian et al.    & 2021          & EM             & \begin{tabular}[c]{@{}l@{}}AutoML platform embedded with many machine learning models is used to \\ detect soil moisture content\end{tabular}                                                                                                                                   & {\cite{babaeian2021estimation}}           \\
Wang et al.        & 2021          & CGSM           & \begin{tabular}[c]{@{}l@{}}UR, MC, PLSR, ANN, RF, and SVM are used to evaluate the nutrient elements \\ in whole rice crop growth stages based on the UAV hyperspectral image\end{tabular}                                                                                      & {\cite{wang2021estimation}}           \\
Zhou et al.        & 2016          & BE             & \begin{tabular}[c]{@{}l@{}}A wheat biomass estimation method based on the combination of HJ-CCD \\ vegetation index and RF is proposed\end{tabular}                                                                                                                             & {\cite{zhou2016estimation}}           \\
Bahrami et al.     & 2021          & BE             & \begin{tabular}[c]{@{}l@{}}SVM, RF, GBDT, XGBoost, and a DNN are used to estimate leaf area index and \\ biomass of three crops based on radar and optical earth detection.\end{tabular}                                                                                          & {\cite{bahrami2021deep}}            \\
Geng et al.        & 2021          & BE             & \begin{tabular}[c]{@{}l@{}}RF, SVM, ANN, and XGBoost are used to estimate seasonal corn biomass based on\\ field observation data and MODIS reflectance data from 2012 to 2019\end{tabular}                                                                                    & {\cite{bahrami2021deep}}          \\
Perez-Ortiz et al. & 2015          & WE             & \begin{tabular}[c]{@{}l@{}}A system is proposed for drawing weed maps using UAVs images which add \\ Hough transform to detect crop row information to improve the accuracy\end{tabular}                                                                                      & {\cite{perez2015semi}}          \\
Reedha et al.      & 2022          & WE             & \begin{tabular}[c]{@{}l@{}}ViT model is used to classify weeds and crops combined with high-resolution \\ images obtained by UAV\end{tabular}                                                                                                                                   & {\cite{reedha2022transformer}}           \\
Alsuwaidi et al.   & 2018          & CDAP           & \begin{tabular}[c]{@{}l@{}}An classification framework which integrates adaptive feature selection, \\ novelty detection and integrated learning is proposed to detect plant diseases \\ and stress conditions and classify crop types based on hyperspectral data\end{tabular} & {\cite{alsuwaidi2018feature}}          \\

Shin et al.        & 2020          & CDAP           & \begin{tabular}[c]{@{}l@{}}Three feature extraction methods (HOG, SURF, and GLCM) and two classifiers\\ (ANN and SVM) are used to detect strawberry powdery mildew\end{tabular}                                                                          & {\cite{shin2020effect}}     \\ \hline
\multicolumn{5}{l}{EM=Environmental monitoring; CGSM=Crop growth state monitoring; BE=Biomass estimation; WE=Weed extraction; CDAP=Crop diseases and pest monitoring.}
\end{tabular}
}
\label{tablegrowth}
\end{table*}

\subsubsection{Weed extraction:}

\begin{figure}[!htbp]
    \begin{minipage}{\linewidth}
        \centering
        \includegraphics[width=1\linewidth]{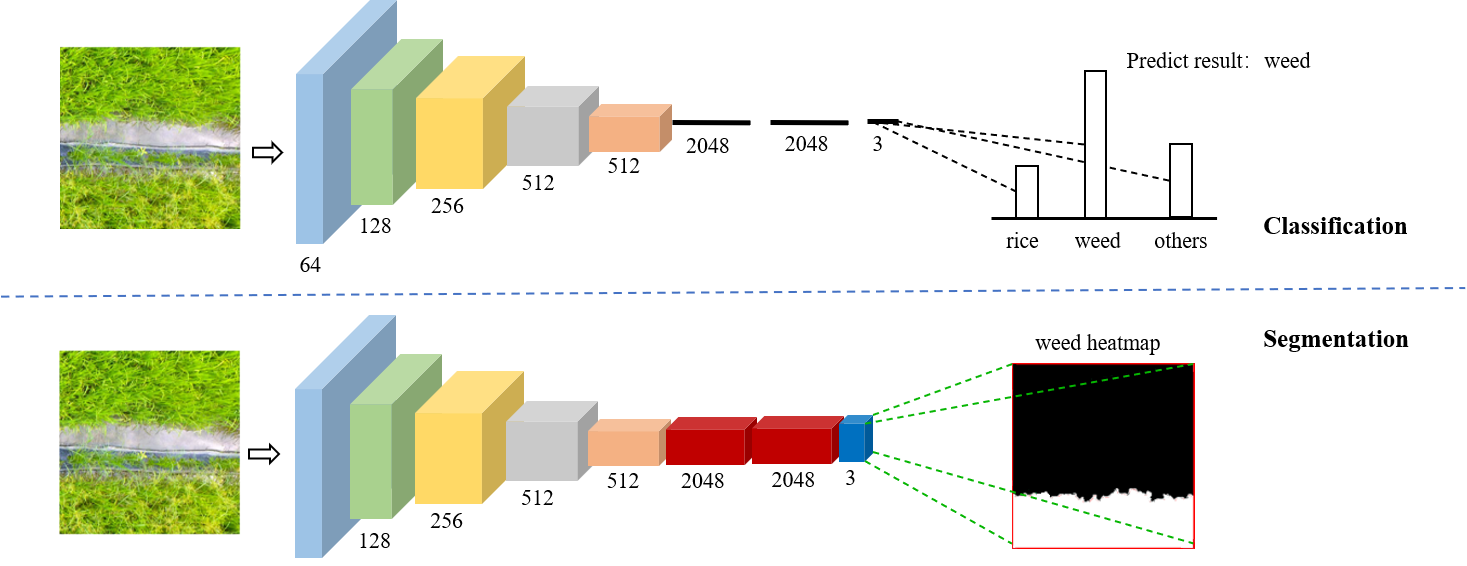}
    \end{minipage}    
    \caption{Comparison of classification and segmentation networks in weed extraction \cite{huang2018fully}.}
    \label{ll_classification and segmentation}
\end{figure}

Fig. \ref{ll_classification and segmentation} presents the deep learning (DL) solutions that separately use classification and segmentation networks for weed detection.

\subsection{Agrifood yield prediction}

Table \ref{tableyield} summarizes some work on crop yield prediction using RS images with a series of artificial intelligence (AI) technologies including linear, non-linear, ensemble ML models and DL networks.

\begin{table}[!ht]
    \centering
    \tiny
    \caption{Summary of yield prediction methods based on RS images and AI technologies.}
    \resizebox{\linewidth}{!}{
    \begin{tabular}{c c c c c c c}
    \bottomrule
        Author & Model & Crop & RS datatype &  RS variable & Period & Metrics \\ \hline
        \cite{johnson2016crop} & MLR, BNN & {\makecell[c]{barley, canola,\\ spring wheat}} & MODIS, AVHRR & NDVI, EVI & 2000-2011 & MAE \\ 
        \cite{fu2020wheat} & {\makecell[c]{LR, MLR, SMLR, \\PLSR, ANN, RF}} & wheat & UAV & LAI, Leaf Dry Matter & 2018–2019 & $\text{R}^{2}$, RRMSE \\ 
        \cite{kamir2020estimating} & {\makecell[c]{RF, XGBoost, Cubist, MLP\\ SVR, GP, KNN, MARS}} & wheat & MOD13Q1 & NDVI & 2009-2015 & $\text{R}^{2}$, RMSE, RMSPE \\ 
        \cite{han2020prediction} & {\makecell[c]{KNNR, NN, DT, SVM, GPR, RF,\\ Boost Trees, Bagging Trees}} & winter wheat & MOD13Q1 & NDVI, EVI & 2001-2014 & $\text{R}^{2}$, RMSE \\ 
        \cite{kayad2019monitoring} & Multiple Regression, RF, SVM & corn & Sentinel-2 & {\makecell[c]{Green Normalized Difference\\ Vegetation Index}} & 2016-2018 & $\text{R}^{2}$, RMSE, MAE, ME, MAPE \\ 
        \cite{zhang2019combining} & LASSO, RF, XGBoost, LSTM & maize & {\makecell[c]{MOD13A2,\\ Contiguous Solar-induced\\ Chlorophyll Fluorescence}} & {\makecell[c]{EVI, SIF}} & 2001-2015 & RMSE \\ 
        \cite{nevavuori2019crop} & 2D-CNN & wheat, barley & UAV & NDVI & 2017 & MAE \\  
        \cite{zhang2021integrating} & LSTM, LASSO, LightGBM & maize & Landsat 5, Landsat 7 & {\makecell[c]{NDVI, EVI, Green Normalized \\Difference
        Vegetation Index, \\Green Chlorophyll\\ Vegetation Index,\\ Wide Dynamic\\ Range Vegetation Index, \\Simple Ratio}} & 2010–2012 & $\text{R}^{2}$, RMSE \\ 
        \cite{jeong2022predicting} & {\makecell[c]{FFNN, 1D-CNN, LSTM,\\ 1D-CNN+LSTM, LSTM+1D-CNN}} & rice & MOD09A1 & {\makecell[c]{NDVI, OSAVI, EVI, \\Renormalized Difference\\ Vegetation Index, \\Modified Triangular\\ Vegetation Index}} & 2011-2017 & $\text{R}^{2}$, RMSE \\ 
        \cite{nevavuori2020crop} & CNN-LSTM, ConvLSTM, 3D-CNN & {\makecell[c]{wheat, barley,\\ oats}} & UAV & --- & 2018  & $\text{R}^{2}$, RMSE, MAPE\\ \bottomrule
\label{tableyield}    
\end{tabular}
}
\end{table}

\begin{figure}[!htbp]
    \centering
    \includegraphics[width=0.5\linewidth]{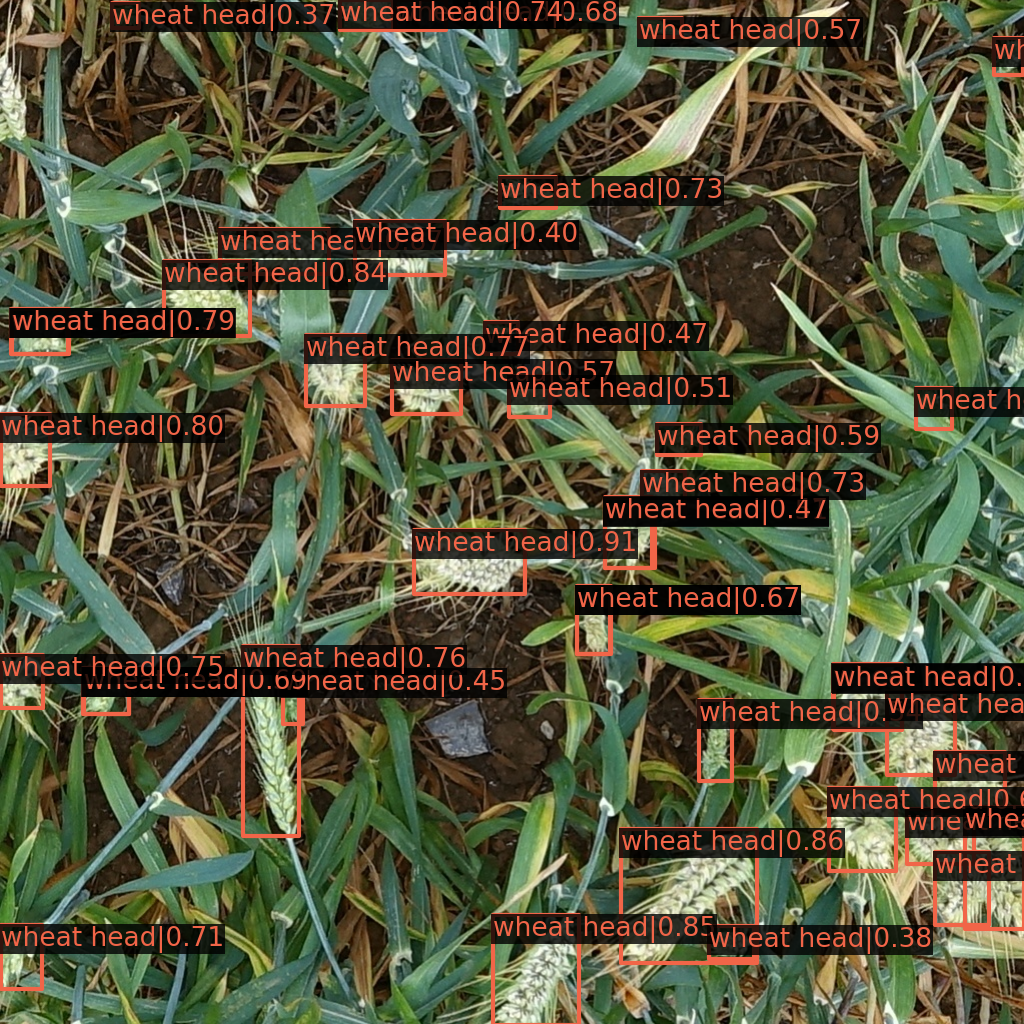}  
    \caption{Detection results of Sparse R-CNN \cite{sparse_r_cnn} using the ViTAEv2-S \cite{zhang2023vitaev2} backbone on the Global Wheat Detection dataset.}
    \label{global_wheat}
\end{figure}

\noindent\textbf{Weed Detection} To show AI technologies have the potential for agrifood yield prediction through natural images, with the Sparse R-CNN detection framework \cite{sparse_r_cnn}, we separately evaluate the performance of different backbones networks, i.e., ResNet-101 \cite{he2016deep} and ViTAEv2-S \cite{zhang2023vitaev2}, on the global wheat detection dataset\footnote{https://www.kaggle.com/competitions/global-wheat-detection/overview}. Experimental results show that, when employing the ResNet-101, although the network has a parameter amount of up to 125.4M, it still performs worse than using the lightweight ViTAE-S, i.e., 41.2 mean average precision (mAP) v.s. 45.2 mAP. Some detection results of ViTAEv2-S are visualized in Fig. \ref{global_wheat}. The number and size of detected wheat instances can be used as indicators for yield prediction.

\section{Artificial Intelligence Methods in Animal Husbandry}

\subsection{Animal behavior monitoring}

\begin{figure}[!tbp]
    \centering
    \includegraphics[width=0.9\linewidth]{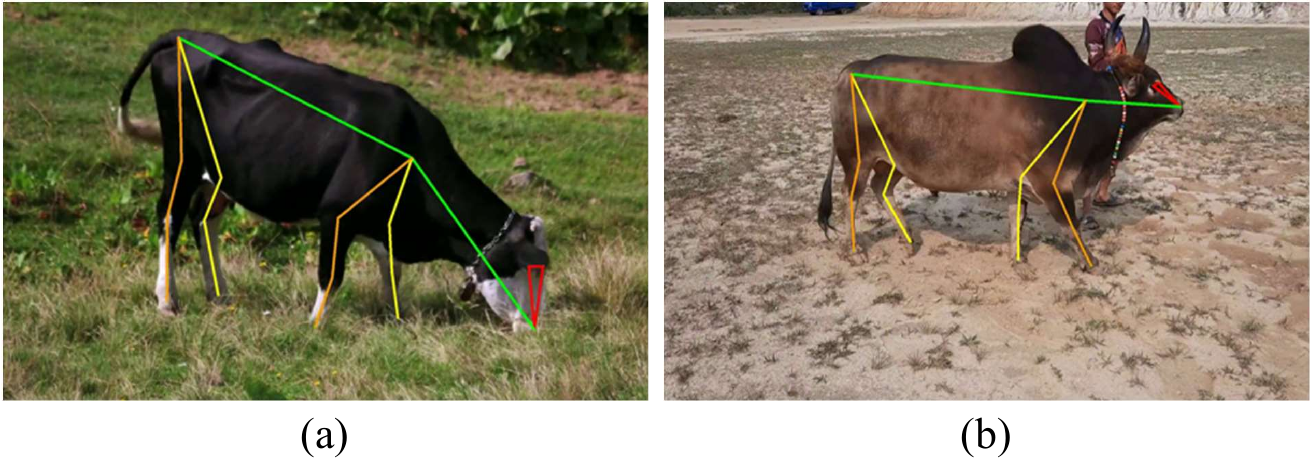}
    \caption{Pose estimation lays the foundation for skeleton-based action recognition and behavior analysis. (a) Eating. (b) Walking.}
    \label{animal_pose}
\end{figure}

In the appendix, we show some pose estimation examples in Fig. \ref{animal_pose}, which are extracted from the AP-10K dataset \cite{yu2ap}. The behaviors of cattle can be recognized through their poses estimated by the ViTPose model \cite{xu2022vitpose}. 

\section{Artificial Intelligence Methods in Fishery}
\subsection{Fishing area identification and prediction}

\par 
In Table~\ref{fishery}, we summarize some typical works on fishery-related tasks with AI algorithms, the key information collected includes Data types, Models/Algorithms, and their Performances/Findings.

\begin{table*}[]
\newcommand{\tabincell}[2]{\begin{tabular}{@{}#1@{}}#2\end{tabular}}
\centering
\caption{Summary of recent methods investigating fishery-related tasks using AI technologies.}
\resizebox{\linewidth}{!}{
\begin{tabular}{cccccc} \hline
& Reference                   & Data type                           & Model/Algorithm                           & Novelty/Targeted problem &         Performance/Finding                   \\ \hline
& \cite{cheng2020research}    & GF-2                                & HDCUNet,FCN-8s,SegNet,U-Net                & \tabincell{c}{Misjudgment of floatage\\ on water surfaces}     &       \tabincell{c}{Overall accuracy (OA) of 99.2\%}            \\
& \cite{zeng2019extracting}   & Landsat TM,OLI,GF-1             & SVM                                        & \tabincell{c}{Water surface geometric features \\ in RS images} &        \tabincell{c}{Distinguish aquaculture ponds\\  from natural water}  \\
& \cite{zeng2020rcsanet}      & Landsat TM/OLI                      & RCSANet                                    & \tabincell{c}{Feature relation between\\ image pixels}  &  \tabincell{c}{OA of 85\% at Lake Hong and \\83\% at Lake Liangzi regions}                \\
& \cite{ottinger2017large}    & Sentinel-1                          & Segmentation algorithm                     & Novel data                       &          \tabincell{c}{OA of 83\% for four coastal \\ aquaculture pond sites}             \\
& \cite{liang2021semi}        & GF-2/GF-1                           & Semi-SSN,FCN8s,Unet,SegNet,HDCUNet         & Difficulty in labeling               &     \tabincell{c}{OA of 90.6\% on the Fujian coastal area}              \\
& \cite{sivasankari2022he}    & NOAA-17,METOP-1,METOP-2             & \makecell{DEEPFISHNETS,BILSTM,SVM,\\GBDT,RF,KNN,NB,ANN} & Hybrid prediction architecture      &     \tabincell{c}{99\% prediction accuracy\\ in the experimental area}         \\
& \cite{shi2018automatic}     & GF-1                                & HCN,DS-HCN,DeepLab                         & Difficulty in labeling          &       \tabincell{c}{Achieving sea–land segmentation and \\ raft labeling in a uniform framework}                 \\
& \cite{knudby2010predictive} & IKONOS                              & LM,GAM,Bagging,RF,Boosted trees,SVM        & Multidimensional variable combination    &   Estimation RMSE of 0.73  \\
& \cite{li2021seecucumbers}   & Drone image                         & YOLOv3                                     & Reduce data volume              &    \tabincell{c}{Detector performance reaches 0.855 mAP, \\ 0.82 precision, 0.83 recall, and 0.82 F1 score}                    \\
& \cite{zhang2022coastal}     & Video                               & Mask R-CNN                                   & Novel dataset                    &        \tabincell{c}{0.628 mAP of Mask R-CNN on the\\ developed dataset}          \\
& \cite{chen2022remote}       & \makecell{MOD09GA,Landsat-7,\\Landsat-8,GSW,JRC} & XGBoost,DNN,RF                             & Multidimensional variable combination           &   \tabincell{c}{The fish catches in Poyang Lake are susceptible \\ to water ecological variables}     \\
& \cite{cao2020real} & Images selected from camera video &  Faster MSSDLite & Real-time detection & \tabincell{c}{AP of 99.01\% and F1 score of 98.94\% with \\ a detection speed of 74.07 frames per second} \\
& \cite{ji2023real} & Images selected from camera video & MobileCenterNet & \tabincell{c}{Underwater crab image enhancement \\ and real-time detection} & \tabincell{c}{AP and F1 values are 97.86\% and 97.94\%, \\and the detection speed is reaching 48.18 frames/s}
\\

& \cite{yao2024fmrft}     & Camera video                  & FMRFT                                        & Fish tracking  & \tabincell{c}{Identification F1 score of 0.903 and \\ multi-object tracking accuracy of 0.943}  \\ 

& \cite{mahmood2020automatic} & Synthetic data & YOLOv3  & Detecting western rock lobsters  & \tabincell{c}{Synthetic data can improve lobster \\detection by a significant margin}\\
& \cite{vo2020application}    & Camera photograph                   & \tabincell{c}{Combining Mask R-CNN and image\\ processing techniques}   &     \tabincell{c}{South Rock Lobster (SRL)\\ biometric recognition}        &   \tabincell{c}{Enable an autonomic grading solution \\ in the SRL supply chain} \\
& \cite{monteiro2023fish}     & Mobile phone photo                  & CNN                                        & Fish species identification  & \tabincell{c}{86\% Accuracy of 8 common fish species \\ in East Coast of Ceará, Brazil}  \\ 
\hline      
\label{fishery} 
\end{tabular}
}
\end{table*}

\section{Opportunities}

\subsection{Foundation Models Contribute Modern Agricultural Systems }

\subsubsection{Large-scale data pre-training}

\begin{figure*}[!htbp]
    \begin{minipage}{1\linewidth}
        \centering
        \includegraphics[width=0.7\linewidth]{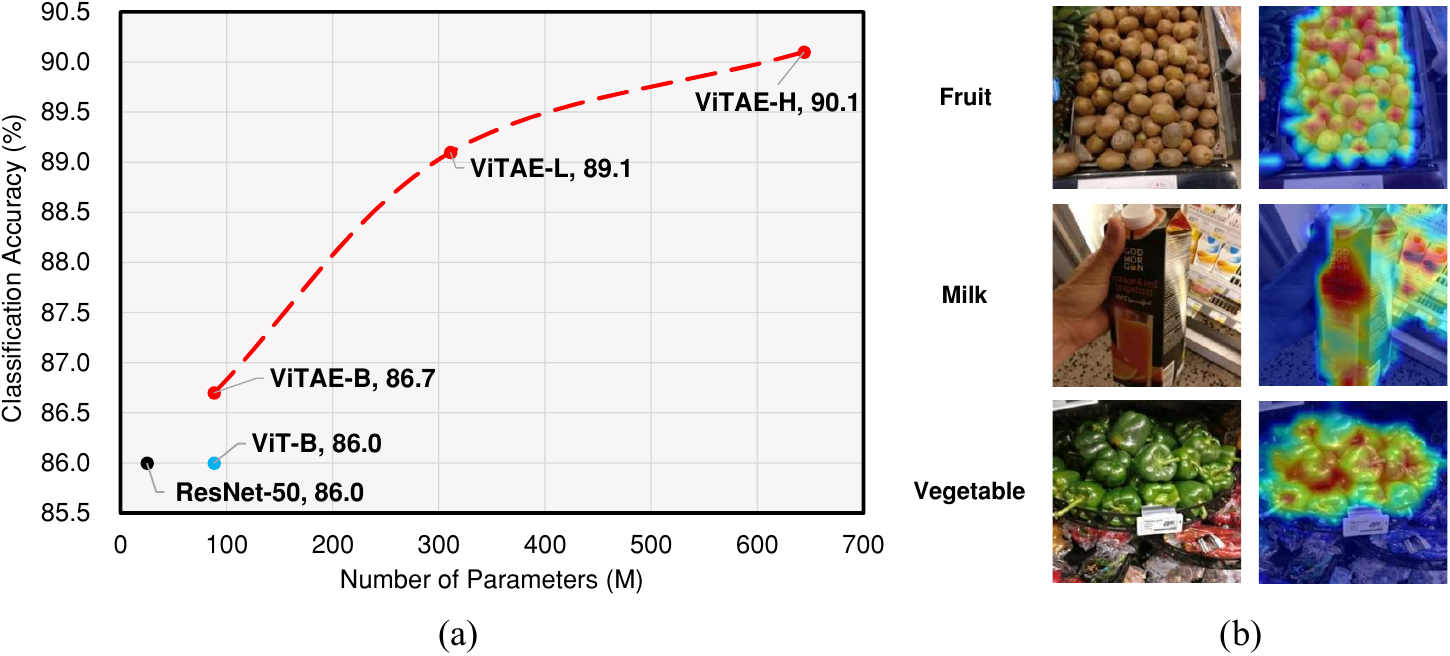}
    \end{minipage}    
    \caption{Performances of large vision models on Grocery Store dataset. (a) Larger ViTAE models deliver higher accuracy than existing classical models. (b) ViTAE model assigns high responsibilities to the region of agrifood.}
    \label{vitae_grocery}
\end{figure*}

 We finetune different ImageNet pre-trained models on the grocery store dataset \cite{grocery_store_dataset} to more intuitively demonstrate the capabilities of large models. Here, we use the MAE-pretrained weights \cite{mae} of the ViTAE series \cite{xu2021vitae,zhang2023vitaev2} from the official website as the model initialization. The ImageNet dataset is used during the pretraining process. Then we follow the settings in ViT \cite{dosovitskiyimage} and Swin \cite{liu2022swin} for the classification task, where an extra classification linear layer is introduced following the backbone to predict the label of the given images. We finetune the whole model on the training set and report the results on the validation set. The dataset split follows the official setting of the Grocery Store dataset. Fig.~\ref{vitae_grocery} (a) depicts that, under the same finetuning setting, the performance of ViTAE-B \cite{xu2021vitae} is superior to classical ViT-B \cite{dosovitskiyimage} and ResNet-50 \cite{he2016deep}, and the accuracy can be further increased when enlarging the model capacity. While the heatmaps in Fig.~\ref{vitae_grocery} (b) indicate the advanced ViTAE model can accurately perceive the agrifood information inside images. 

\subsection{Trustworthy AI Technologies Rebuild Our Agrifood systems}

\subsubsection{Safe and traceable technology}

\begin{figure*}[!htbp]
    \begin{minipage}{1\linewidth}
        \centering
        \includegraphics[width=\linewidth]{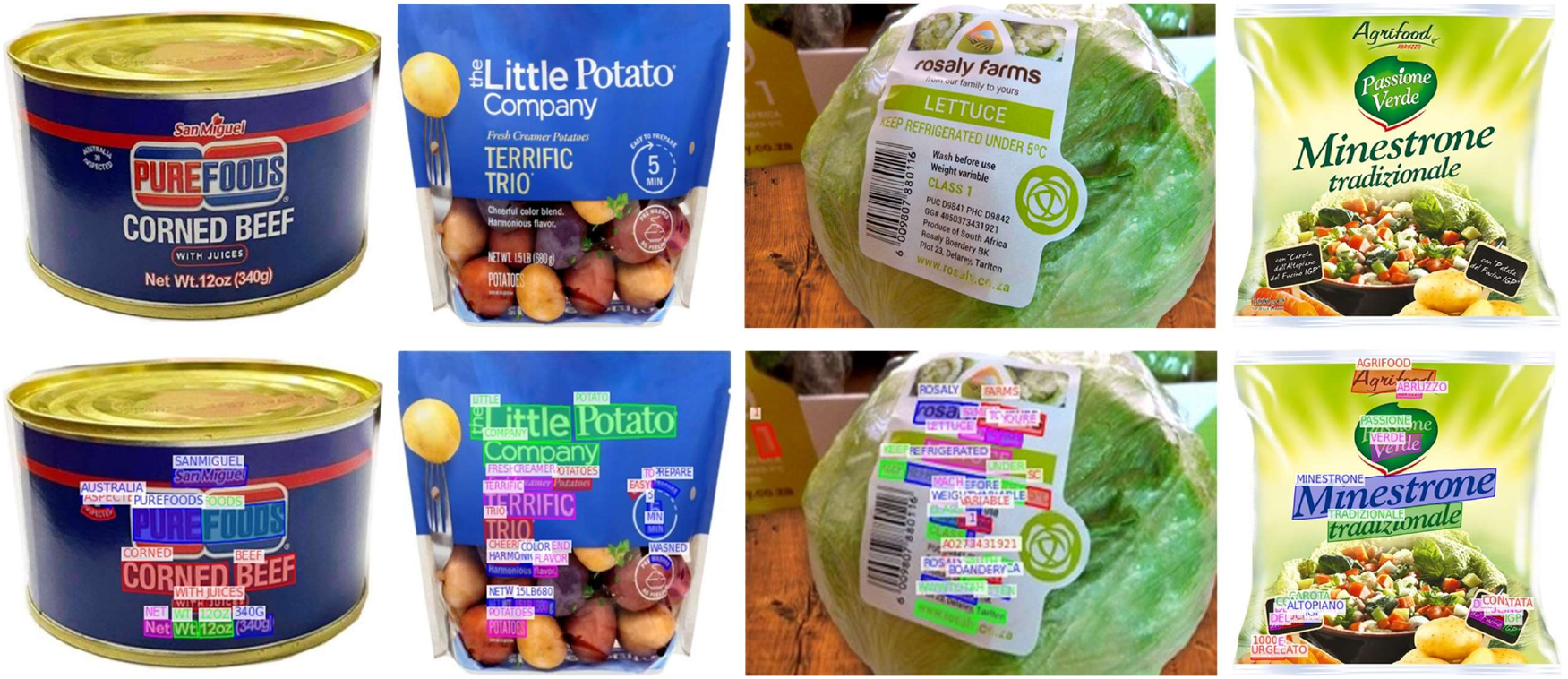}
    \end{minipage}    
    \caption{AI models enable reading agrifood product information. The results are obtained by DeepSolo~\cite{ye2023deepsolo} with ViTAEv2-S~\cite{zhang2023vitaev2}.}
    \label{ar_text_detect}
\end{figure*}

Fig. \ref{ar_text_detect} displays some text spotting examples to show how AI technology enables reading the product
information of agrifood.

\end{document}